\documentclass[letterpaper]{article} 
\usepackage{aaai24}  
\usepackage{times}  
\usepackage{helvet}  
\usepackage{courier}  
\usepackage[hyphens]{url}  
\usepackage{graphicx} 
\usepackage{natbib}  
\usepackage{caption} 
\usepackage{algorithm}
\usepackage{algorithmic}

\usepackage{newfloat}
\usepackage{listings}
\DeclareCaptionStyle{ruled}{labelfont=normalfont,labelsep=colon,strut=off} 
\floatstyle{ruled}
\newfloat{listing}{tb}{lst}{}
\floatname{listing}{Listing}
\def\ie{{\it{i.e.,~}}}

\usepackage{makecell}
\usepackage{subcaption}
\usepackage{booktabs}       
\usepackage{multirow}
\usepackage{amssymb}
\usepackage{amsmath}
\usepackage[dvipsnames, svgnames, x11names]{xcolor}
\usepackage{colortbl}

\usepackage{cleveref}

\title{SCTNet: Single-Branch CNN with Transformer Semantic Information for Real-Time Segmentation}
\author{
    Zhengze Xu\textsuperscript{\rm 1}\thanks{Work strenthened during an internship at Meituan.}, Dongyue Wu\textsuperscript{\rm 1}, Changqian Yu\textsuperscript{\rm 2}, Xiangxiang Chu\textsuperscript{\rm 2}, Nong Sang\textsuperscript{\rm 1}, Changxin Gao\textsuperscript{\rm 1}\thanks{Corresponding author}
}
\affiliations{
    \textsuperscript{\rm 1}National Key Laboratory of Multispectral Information Intelligent Processing Technology, School of Artificial Intelligence and Automation, Huazhong University of Science and Technology\\
    \textsuperscript{\rm 2}Meituan\\

    \{zhengzexu, dongyue\_wu\, nsang, cgao\}@hust.edu.cn, y-changqian@outlook.com, cxxgtxy@gmail.com
}

\usepackage{bibentry}

\begin{document}

\maketitle

\begin{abstract}
Recent real-time semantic segmentation methods usually adopt an additional semantic branch to pursue rich long-range context. However, the additional branch incurs undesirable computational overhead and slows inference speed. 
To eliminate this dilemma, we propose SCTNet, a single branch CNN with transformer semantic information for real-time segmentation. SCTNet enjoys the rich semantic representations of an inference-free semantic branch while retaining the high efficiency of lightweight single branch CNN.
SCTNet utilizes a transformer as the training-only semantic branch considering its superb ability to extract long-range context. With the help of the proposed transformer-like CNN block CFBlock and the semantic information alignment module, SCTNet could capture the rich semantic information from the transformer branch in training. During the inference, only the single branch CNN needs to be deployed. We conduct extensive experiments on Cityscapes, ADE20K, and COCO-Stuff-10K, and the results show that our method achieves the new state-of-the-art performance. The code and model is available at https://github.com/xzz777/SCTNet.
\end{abstract}

\section{Introduction}
As a fundamental task in computer vision, semantic segmentation aims to assign a semantic class label to each pixel in the input image. It plays a vital role in autonomous driving, medical image processing, mobile applications, and many other fields. In order to achieve better segmentation performance, recent semantic segmentation methods pursue abundant long-range context. Different methods have been proposed to capture and encode rich contextual information, including large receptive fields~\cite{chen2014semantic,chen2017deeplab,chen2018encoder}, multi-scale feature fusion~\cite{ronneberger2015u,zhao2017pyramid}, self-attention mechanism~\cite{fu2019dual,huang2019ccnet,yuan2018ocnet,zhao2018psanet,dosovitskiy2020image}, etc. Among them, the self-attention mechanism, as an essential component of transformers, has been proven to have a remarkable ability to model long-range context. Although these works improve significantly, they usually lead to high computational costs. Note that self-attention-based works even have square computation complexity with respect to image resolution, which significantly increases latency in processing high-resolution images. These limitations hinder their application in real-time semantic segmentation.

\begin{figure}[tb]
  \begin{center}
     \includegraphics[width=\linewidth]{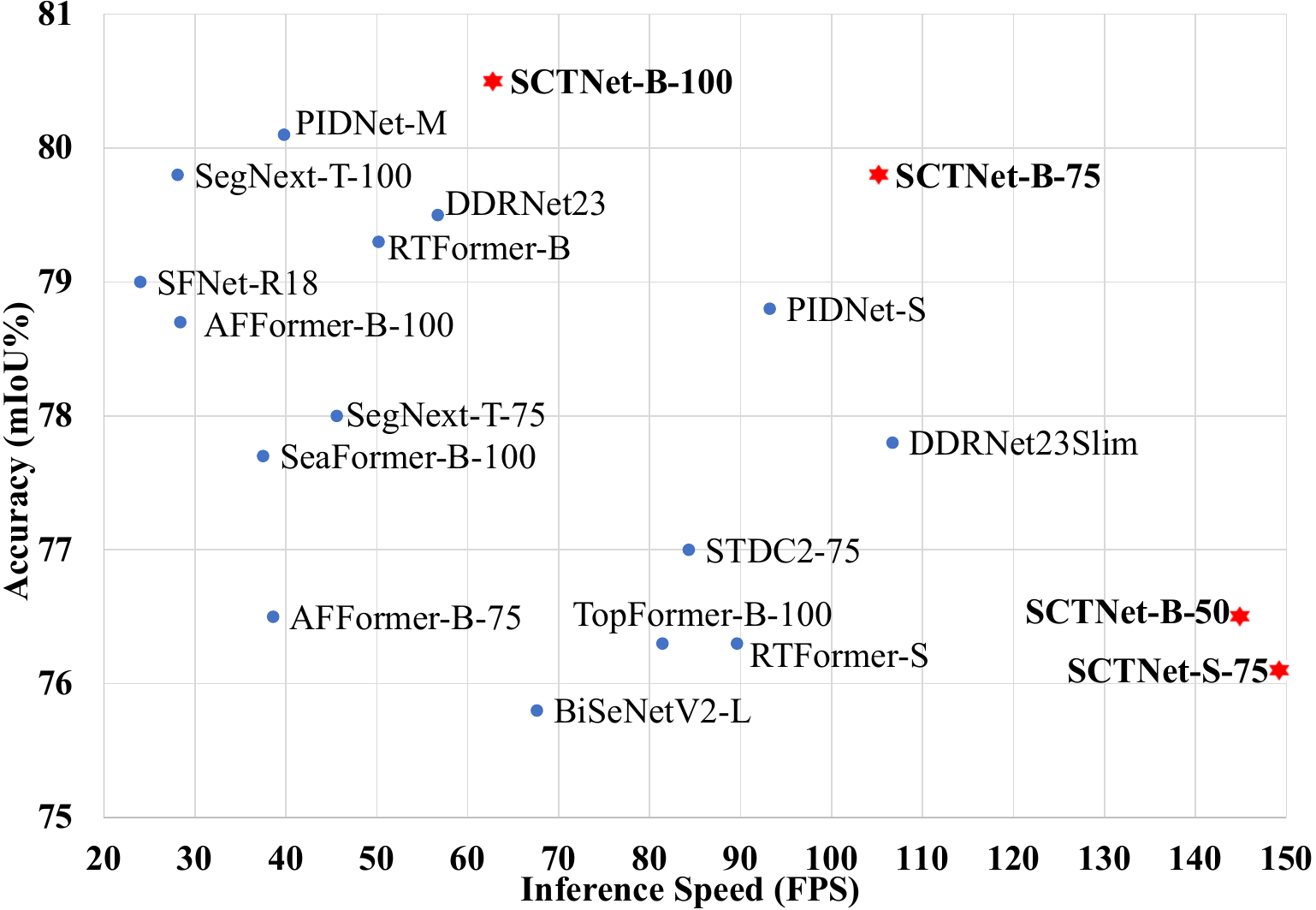}
  \end{center}
  \vspace{-10pt}
     \caption{\textbf{The speed-accuracy performance on Cityscapes validation set}. Our methods are presented in red stars, while others are presented in blue dots. Our SCTNet establishes a new state-of-the-art speed-accuracy trade-off.}
  \label{fig:figure1_trade-off}
  \vspace{-10pt}
  \end{figure}

\begin{figure}
  \begin{center}
     \includegraphics[width=\linewidth]{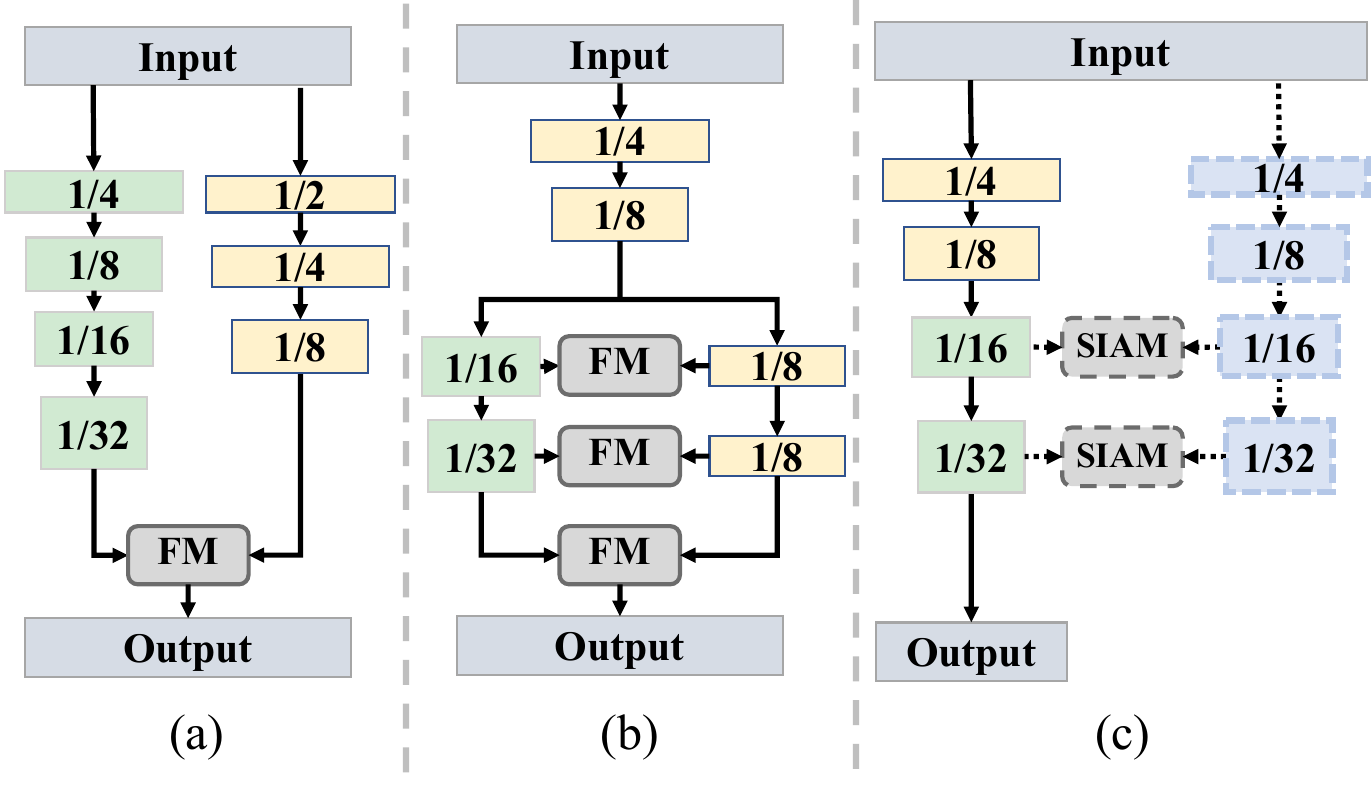}
  \setlength{\belowdisplayskip}{-10pt}
  \end{center}
  \vspace{-10pt}
     \caption{\textbf{Real-time semantic segmentation paradigms.} (a) \emph{Decoupled bilateral network} divides a semantic branch and a spatial branch at the early stage. (b) \emph{Feature sharing bilateral network} separates the two branches at the latter stage and adopts dense fusion modules. (c) Our \textbf{SCTNet} applies a single hierarchy branch with a semantic extraction transformer, free from the extra branch and costly fusion module in inference. \textbf{FM}: Fusion Module,  \textbf{SIAM}: Semantic Information Alignment Module. Dashed arrows and boxes denote training-only. }
  \label{fig:figure2_comparsion}
  \vspace{-10pt}
  \end{figure}

Many recent real-time works adopt a bilateral architecture to extract high-quality semantic information at a fast speed. BiSeNet~\cite{yu2018bisenet} proposes a bilateral network to separate the detailed spatial features and ample contextual information at early stages and process them in parallel, which is shown in Figure~\ref{fig:figure2_comparsion}(a). Following BiseNet~\cite{yu2018bisenet}, BiSeNetV2~\cite{yu2021bisenet} and STDC~\cite{fan2021rethinking} make further efforts to strengthen the capability to extract rich long-range context or reduce the computational costs of the spatial branch. To balance inference speed and accuracy, DDRNet~\cite{pan2022deep}, RTFormer~\cite{wang2022rtformer}, and SeaFormer~\cite{wan2023seaformer} adopt a feature-sharing architecture that divides spatial and contextual features at the deep stages, as shown in Figure~\ref{fig:figure2_comparsion}(b). However, these methods introduce dense fusion modules between two branches to boost the semantic information of extracted features. In conclusion, all these bilateral methods suffer from limited inference speed and high computational costs due to the additional branch and multiple fusion modules.

To eliminate the aforementioned dilemma, we propose a single-branch CNN with transformer semantic information for real-time segmentation~(SCTNet). It can extract semantic information efficiently without heavy computation caused by the bilateral network. Specifically, SCTNet learns long-range context from a training-only transformer semantic branch to the CNN branch. To mitigate the semantic gap between the transformer and CNN, we elaborately design a transformer-like CNN block called CFBlock and utilize a shared decoder head before the alignment. With the aligned semantic information in training, the single-branch CNN can encode the semantic information and spatial details jointly. Therefore, SCTNet could align the semantic representation from the large effective receptive field of transformer architecture while maintaining the high efficiency of a lightweight single branch CNN architecture in inference. The overall architecture is illustrated in Figure~\ref{fig:figure2_comparsion}(c). Extensive experimental results on three challenging datasets demonstrate that the proposed SCTNet has a better trade-off between accuracy and speed than previous works.
Figure~\ref{fig:figure1_trade-off} intuitively shows the comparison between SCTNet and other real-time segmentation methods on the Cityscapes val set.

The main contributions of the proposed SCTNet can be summarized as the following three aspects:
\begin{itemize}
    \item We propose a novel single-branch real-time segmentation network called \textbf{SCTNet}. By learning to extract rich semantic information utilizing semantic information alignment from the transformer to CNN, SCTNet enjoys \textbf{high accuracy of the transformer} while maintaining \textbf{fast inference speed of the lightweight single branch CNN.}
    \item To alleviate the semantic gap between CNN features and transformer features, we design the \textbf{CFBlock} (Conv-Former Block), which could capture long-range context as a transformer block using only convolution operations. Moreover, we propose \textbf{SIAM}(Semantic Information Alignment Module) to align the features in a more effective way.
    \item Extensive experimental results show that the proposed SCTNet \textbf{outperforms existing state-of-the-art} methods for real-time semantic segmentation on Cityscapes, ADE20K, and COCO-Stuff-10K. SCTNet provides a new view of boosting the speed and improving the performance for real-time semantic segmentation.
\end{itemize}
\section{Related Work}
\textbf{Semantic Segmentation.} 
FCN~\cite{long2015fully} leads to the tendency to utilize CNN for semantic segmentation. Following FCN, a series of improved CNN-based semantic segmentation methods are proposed. DeepLab~\cite{chen2017deeplab} enlarges the receptive field with dilated convolution. PSPNet~\cite{zhao2017pyramid}, U-Net~\cite{ronneberger2015u}, and RefineNet~\cite{lin2017refinenet} fuse different level feature representations to capture multi-scale context. Some methods~\cite{fu2019dual,huang2019ccnet,yuan2018ocnet,zhao2018psanet}propose various attention modules to improve segmentation performance. In recent years, transformer has been adopted for semantic segmentation and shows promising performance. SETR~\cite{zheng2021rethinking} directly applies the vision transformer to image segmentation for the first time. SegViT~\cite{zhang2022segvit} tries to perform semantic segmentation with the plain vision transformer. PVT~\cite{wang2021pyramid} introduces the typical hierarchical architecture in CNN into the transformer-based semantic segmentation model. SegFormer~\cite{xie2021segformer} proposes an efficient multi-scale transformer-based segmentation model.

\begin{figure*}
\begin{center}
   \includegraphics[width=\linewidth]{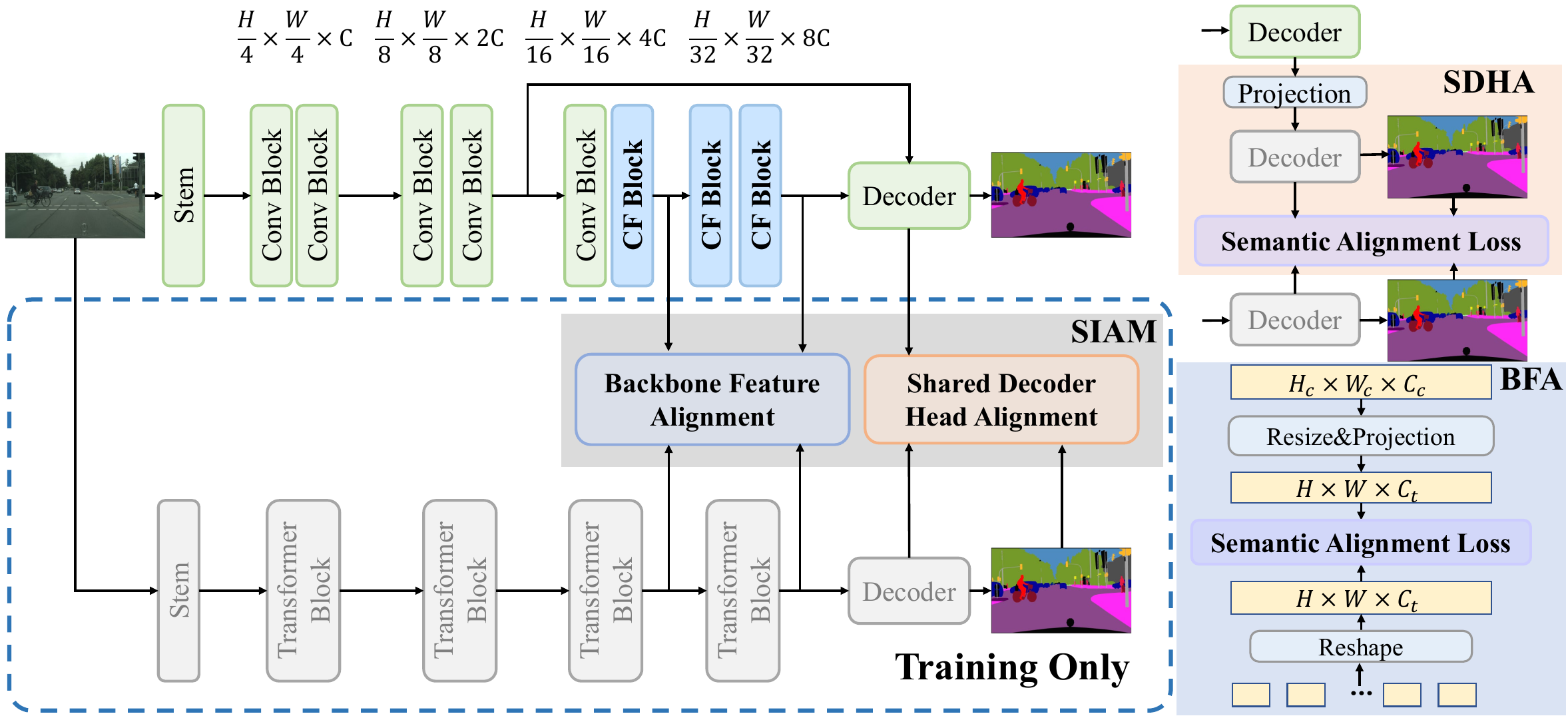}
\end{center}
   \caption{\textbf{The architecture of SCTNet}. CFBlock (Conv-Former Block, detailed in Figure~\ref{fig:figure4_ConvFormerBlock}) takes advantage of the training-only Transformer branch (greyed-out in the dashed box) via SIAM (Semantic Information Alignment Module) which is composed of BFA (Backbone Feature Alignment) and SDHA (Shared Decoder Head Alignment).}
\label{fig:figure3_OverallArchitecture}
\vspace{-10pt}
\end{figure*}

\noindent
\textbf{Real-time Semantic Segmentation.} 
Early real-time semantic segmentation methods~\cite{paszke2016enet,wu2017real} usually accelerate inference by compressing channels or fast down-sampling. ICNet~\cite{zhao2018icnet} first introduces a multi-resolution image cascade network to accelerate the speed. BiSeNetV1~\cite{yu2018bisenet} and BiSeNetV2~\cite{yu2021bisenet} adopt two-branch architecture and feature fusion modules to achieve a better trade-off between speed and accuracy. STDC~\cite{fan2021rethinking} rethinks the two-branch network of BiSeNet, removes the spatial branch, and adds a detailed guidance module. DDRNets~\cite{pan2022deep} achieves a better trade-off by sharing branches in the early stages. Very recently, some efficient transformer methods for real-time segmentation have been proposed, but they still have unresolved problems. TopFormer~\cite{zhang2022topformer} only uses transformer on 1/64 scale of the feature maps, leading to low accuracy. RTFormer~\cite{wang2022rtformer} and SeaFormer~\cite{wan2023seaformer} need frequent interaction between the two branches. This additional computation slows down the inference speed. There are also some single-branch and multi-branch methods. See more discussions in appendix~\ref{section:D}.

\noindent
\textbf{Attention mechanism.} 
Attention mechanism has been widely used in computer vision in recent years. Many methods~\cite{chu2021twins,liu2021swin,fang2022msg} contribute to reducing the computational complexity of attention mechanisms. Although some of them have achieved linear complexity, they contain frequent shift or reshape operations which bring lots of latency. MSCA~\cite{guo2022segnext} shows a promising performance, but the large kernel is not employ-friendly, and the multi-scale design of attention further incurs inference speed. External attention~\cite{guo2022beyond} has a very simple form. It uses external parameters as the key and value and implements the attention mechanism with two linear layers. GFA(GPU-Friendly Attention)~\cite{wang2022rtformer} improves external attention by replacing head split in EA with group double norm, which is more friendly for GPU devices.


\section{Methodology}

\subsection{Motivation}

Removing the semantic branch of bilateral networks can significantly speed up the inference. However, this results in shallow single-branch networks that lack long-range semantic information, leading to low accuracy. While using deep encoders and powerful decoders or complex enhancement modules can recover accuracy, it slows down the inference process. To address this issue, we propose a training-only alignment method that enriches semantic information without sacrificing inference speed. Specifically, we proposed SCTNet, a single-branch convolution network with a training-only semantic extraction transformer, which owns high accuracy of transformer and fast inference speed of CNN. The overview of SCTNet is presented in Figure~\ref{fig:figure3_OverallArchitecture}.

\subsection{Conv-Former Block}
As different types of networks, the feature representations extracted by CNN and transformer significantly differ. Directly aligning the features between the CNN and the transformer makes the learning process difficult, resulting in limited performance improvement. In order to make the CNN branch easily learns how to extract high-quality semantic information from the transformer branch, we design the Conv-Former Block. Conv-Former Block simulates the structure of the transformer block as much as possible to learn the semantic information of the transformer branch better. Meanwhile, the Conv-Former Block implements the attention function using only efficient convolution operations.

\begin{figure}
  \begin{center}
     \includegraphics[width=\linewidth]{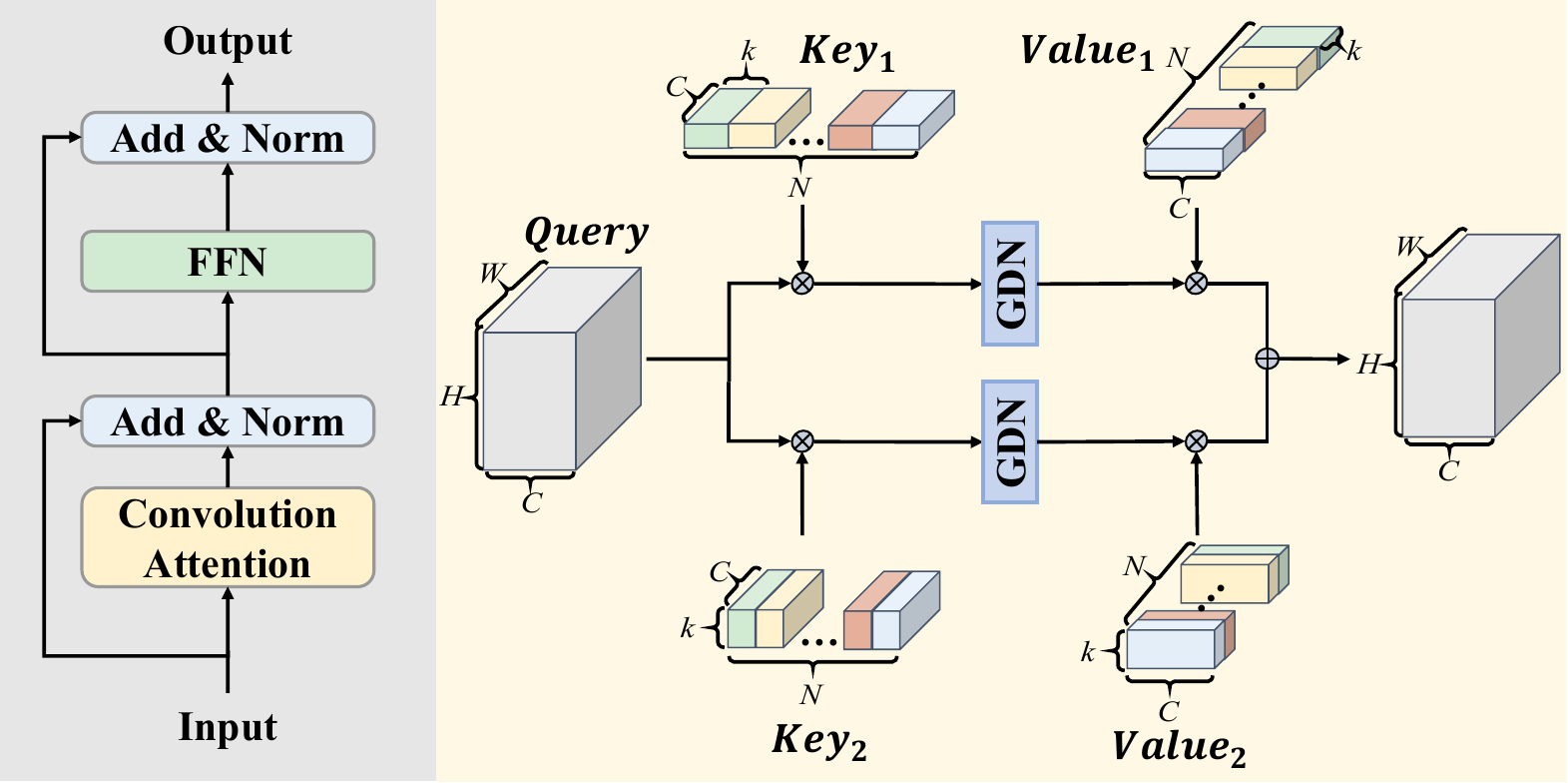}
  \end{center}
     \caption{Design of Conv-Former Block (left) and the details of convolutional attention (right). GDN means Grouped Double Normalization. $\otimes$ means convolution operations, $\oplus$ stands for addition, and $k$ means the kernel size.}
  \label{fig:figure4_ConvFormerBlock}
  \vspace{-10pt}
  \end{figure}

The structure of the Conv-Former Block is similar to the structure of a typical transformer encoder~\cite{vaswani2017attention}, as presented in the left of Figure~\ref{fig:figure4_ConvFormerBlock}. 
The process can be described as follows:
\begin{equation}
\label{Equ:ConvFormerBlock}
\begin{aligned}
f &= Norm(x + ConvAttention(x)),\\
y &= Norm (f + FFN(f)),
\end{aligned}
\end{equation}
where $Norm(\cdot)$ refers to  batch normalization~\cite{ioffe2015batch}, and $x$, $f$, $y$ denote input, hidden feature and output, respectively. 

\noindent
{\bf Convolutional Attention.} 
Attention mechanisms used for real-time segmentation should have the property of low latency and powerful semantic extraction ability. As discussed in the related work, We believe GFA is a potential candidate. Our convolutional attention is derived from GFA. 

There are two main differences between GFA and the proposed convolutional attention.  
Firstly, we replace the matrix multiplication in GFA with pixel-wise convolution operations. Point convolution is equivalent to pixel-to-pixel multiplication but without feature flattening and reshaping operations. These operations are detrimental to maintaining the inherent spatial structure and bring in extra inference latency. Moreover, convolution provides a more flexible way to extend external parameters.
Then, due to the semantic gap between the transformer and CNN, it is not enough to capture rich context that simply calculates the similarity between several learnable vectors and each pixel and then enhances the pixels according to the similarity map and the learnable vectors. To better align the semantic information of the transformer, we enlarge the learnable vectors to learnable kernels. On the one hand, this converts the similarity calculation between pixel and learnable vectors to that between pixel patches with learnable kernels. On the other hand, the convolution operation with learnable kernels retains more local spatial information to some extent. 
The operations of convolution attention can be summarized as follows:
\begin{equation}\label{Eq:Conv-Former Attention}
X = \theta \left(X \otimes K\right) \otimes K^{T}, 
\end{equation}
where $\mathbf{X}\in\mathbb{R}^{C \times H \times W}$,$K \in \mathbb{R}^{C \times N \times k \times k}$, $K^{T} \in \mathbb{R}^{N \times C \times k \times k}$ represents input image and learnable query and key, respectively. $C$, $H$, $W$ denote the channel, height, and width of the feature map, respectively. $N$ denotes the number of learnable parameters, and $k$ denotes the kernel size of the learnable parameters. $\theta$ symbolizes the grouped double normalization, which applies softmax on the dimension of $H \times W$ and grouped $L2$ Norm on the dimension of $N$. $\otimes$ means convolution operations.

Taking efficiency into consideration, we implement the convolution attention with stripe convolution rather than standard convolutions. More specifically, we utilize a $1 \times k$ and a $k \times 1$ convolution to approximate a $k \times k$ convolution layer. Figure~\ref{fig:figure4_ConvFormerBlock} illustrate the implementation details of convolution attention. 

\noindent
{\bf Feed Forward Network.} 
Typical FFN plays a vital role in providing position encoding and embedding channels. The typical FFN (Feed Forward Network) in recent transformer models consists of a expand point convolution, a depth-wise $3 \times 3$ convolution, and a squeeze point convolution. Different from typical FFN, our FFN is made up of two standard $3 \times 3$ convolution layers. Compared with the typical FFN, our FFN is more efficient and provides a larger receptive field.

\begin{table*}[t]
    \centering
  \begin{tabular}{c l|c|c|c|c|c}
    \Xhline{1pt}
    Method & Reference &\#Params$\downarrow$ &Resolution &FPS(TRT)$\uparrow$ &FPS(Torch)$\uparrow$ &mIoU(\%)$\uparrow$ \\
    \hline
    SFNet-ResNet18  &ECCV~\citeyear{li2020semantic} &12.3M  &$2048 \times 1024$ &50.5 &24.0 &79.0  \\
    AFFormer-B-Seg100 &AAAI~\citeyear{dong2023afformer} &3.0M &$2048 \times 1024$ &58.3 &28.4 &78.7 \\
    AFFormer-B-Seg75  &AAAI~\citeyear{dong2023afformer} &3.0M &$1536 \times 768$ &96.4 &38.6 &76.5 \\
    AFFormer-B-Seg50  &AAAI~\citeyear{dong2023afformer} &3.0M &$1024 \times 512$ &148.4 &49.5 &73.5 \\
    SegNext-T-Seg100 &NeurIPS~\citeyear{guo2022segnext} &4.3M &$2048 \times 1024$ &46.5 &28.1 &79.8 \\
    SegNext-T-Seg75  &NeurIPS~\citeyear{guo2022segnext} &4.3M &$1536 \times 768$ &78.3 &45.6 &78.0 \\
    PIDNet-S  &CVPR~\citeyear{xu2023pidnet}  &7.6M &$2048 \times 1024$ &127.1 &93.2 &78.8 \\   
    PIDNet-M  &CVPR~\citeyear{xu2023pidnet}  &34.4M &$2048 \times 1024$ &90.7 &39.8 &\textbf{80.1} \\    
    \hline
    \multicolumn{2}{c|}{\textcolor{gray}{CNN-based Bilateral Networks}} & &  & & &\\
    BiSeNet-ResNet18  &ECCV~\citeyear{yu2018bisenet}  &49.0M  &$1536 \times 768$     &182.9 &112.3 &74.8  \\
    BiSeNetV2-L      &IJCV~\citeyear{yu2021bisenet}   &-      &$1024 \times 512$    &102.3 &67.6    &75.8   \\
    STDC1-Seg75 &CVPR~\citeyear{fan2021rethinking} &14.2M &$1536 \times 768$ &209.5 &101.9 &74.5 \\
    STDC2-Seg75 &CVPR~\citeyear{fan2021rethinking} &22.2M &$1536 \times 768$ &149.2 &84.3  &77.0 \\
    STDC1-Seg50 &CVPR~\citeyear{fan2021rethinking} &14.2M &$1024 \times 512$ &397.6 &146.2 &72.2 \\
    STDC2-Seg50 &CVPR~\citeyear{fan2021rethinking} &22.2M &$1024 \times 512$ &279.7 &94.6  &74.2 \\
    DDRNet-23-S &TIP~\citeyear{pan2022deep} &5.7M &$2048 \times 1024$ &138.9 &106.7 &77.8 \\
    DDRNet-23 &TIP~\citeyear{pan2022deep} &20.1M &$2048 \times 1024$ &101.9 &56.7 &\textbf{79.5} \\
    \hline
    \multicolumn{2}{c|}{\textcolor{gray}{Transformer-based Bilateral Networks}} & &  & & &\\
    TopFormer-B-Seg100 &CVPR~\citeyear{zhang2022topformer}    &5.1M &$2048 \times 1024$ &128.4 &81.4 &76.3 \\
    TopFormer-B-Seg50  &CVPR~\citeyear{zhang2022topformer}    &5.1M &$1024 \times 512$ &410.9 &95.7 &70.7\\
    SeaFormer-B-Seg100 &ICLR~\citeyear{wan2023seaformer}    &8.6M &$2048 \times 1024$ &103.6 &37.5 &77.7 \\
    SeaFormer-B-Seg50 &ICLR~\citeyear{wan2023seaformer} &8.6M &$1024 \times 512$ &231.6 &45.2 &72.2 \\
    RTFormer-S &NeurIPS~\citeyear{wang2022rtformer}    &4.8M &$2048 \times 1024$ &- &89.6 &76.3 \\
    RTFormer-B &NeurIPS~\citeyear{wang2022rtformer}    &16.8M &$2048 \times 1024$ &- &50.2 &\textbf{79.3} \\
    \hline
    \textbf{SCTNet-S-Seg50}  &Ours &4.6M &$1024 \times 512$ &\textbf{451.2} &\textbf{160.3} &72.8\\
    \textbf{SCTNet-S-Seg75} &Ours &4.6M &$1536 \times 768$ &233.3 &149.2 &76.1 \\
    \textbf{SCTNet-B-Seg50}  &Ours &17.4M &$1024 \times 512$ &374.6 &144.9 &76.5 \\
    \textbf{SCTNet-B-Seg75}  &Ours &17.4M &$1536 \times 768$ &186.6 &105.2 &79.8\\
    \textbf{SCTNet-B-Seg100} &Ours &17.4M &$2048 \times 1024$ &105.0  &62.8 &\textbf{80.5} \\
    \Xhline{1pt}
  \end{tabular}
\caption{\textbf{Comparisons with other state-of-the-art real-time methods on Cityscapes val set.} Seg100, Seg75, Seg50 denote the input size of $1024\times2048$, $768\times1536$, $512\times1024$, respectively. \#Params refers to the number of parameters.
}
  \label{tab:sample-tableSOTA}
\vspace{-10pt}
\end{table*}

\subsection{Semantic Information Alignment Module}
A simple yet effective alignment module is proposed to conduct the feature learning in the training process, as shown in Figure~\ref{fig:figure3_OverallArchitecture}. It can be divided into backbone feature alignment and shared decoder head alignment.

\noindent
{\bf Backbone Feature Alignment.} 
Thanks to the transformer-like architecture of the Conv-Former Block, the alignment loss can easily align the Conv-Former Block's features with the features of transformers. In short, the backbone feature alignment first down-sample or up-sample the feature from the transformer and CNN branches for alignment. Then it projects the feature of the CNN to the dimension of the transformer. The projection can: $1)$ unify the number of channels and $2)$ avoid direct alignment of features, which damages the supervision of ground truth for the CNN in the training process. Finally, a semantic alignment loss is applied to the projected features to align the semantic representations.

\noindent
{\bf Shared Decoder Head Alignment.} 
Transformer decoders often use the features of multiple stages for complex decoding, while SCTNet decoder only picks the features of stage2\&stage4. Considering the significant difference in decoding space between them, direct alignment of the decoding features and output logits can only get limited improvement. Therefore, we propose shared decoder head alignment. Specifically, the concatenation stage2\&stage4 features of the single-branch CNN are input into a point convolution to expand the dimension. Then the high-dimension features are passed through the transformer decoder. The transformer decoder's new output features and logits are used to calculate alignment loss with the origin outputs of the transformer decoder.

\subsection{Overall Architecture}
To reduce computational costs while obtain rich semantic information, we simplify the popular two-branches architecture to one swift CNN branch for inference and a transformer branch for semantic alignment only for training.

\noindent
{\bf Backbone.} 
To improve the inference speed, SCTNet adopts a typical hierarchical CNN backbone. SCTNet starts from a stem block consisting of two sequential 3$\times$3 convolution layers. The former two stages consist of stacked residual blocks~\cite{he2016deep}, and the latter two stages include the proposed transformer-like blocks called Conv-Former Blocks~(CFBlocks). The CFBlock employs several elaborately designed convolution operations to perform the similar long-range context capturing function of the transformer block. We apply a convdown layer consisting of a stridden convolution with batch normal and ReLu activation for down-sampling at the beginning of stage $2\sim4$, which is omitted in Figure~\ref{fig:figure3_OverallArchitecture} for clarity. 

\noindent
{\bf Decoder Head.} The decoder head consists of a DAPPM~\cite{pan2022deep} and a segmentation head. To further enrich the context information, we add a DAPPM after the output of stage 4. Then we concatenate the output with the feature map of Stage 2. Finally, this output feature is passed into a segmentation head. Precisely, the segmentation head consists of a 3$\times$3 Conv-BN-ReLU operator followed by a 1$\times$1 convolution classifier.

\noindent
{\bf Training Phase.} 
It is well known that transformer excels at capturing global semantic context. On the other hand, CNN has been widely proven to be better at modeling hierarchical locality information than transformers. Motivated by the advantages of transformer and CNN, we explore equipping a real-time segmentation network with both merits. We propose a single-branch CNN that learns to align its features with those of a powerful transformer, which is illustrated in the blue dotted box in Figure~\ref{fig:figure3_OverallArchitecture}. This feature alignment enables the single-branch CNN to extract both rich global context and detailed spatial information. Specifically, there are two streams in the training phase. SCTNet adopts a train-only transformer as the semantic branch to extract powerful global semantic context. The semantic information alignment module supervises the convolution branch to align high-quality global context from the transformer.

\noindent
{\bf Inference Phase.} 
To avoid the sizeable computation costs of two branches, only the CNN branch is deployed in the inference. With the transformer-aligned semantic information, the single-branch CNN can generate accurate segmentation results without the extra semantic extraction or costly dense fusion. Specifically, the input image is fed into a single-branch hierarchy convolution backbone. Then the decoder head picks up the features in the backbone and conducts simple concatenation followed by pixel-wise classification. 

\vspace{-4pt}
\noindent
\subsection{Alignment Loss} 
For better alignment of semantic information, a alignment loss focusing on semantic information rather than spatial information is needed. In the implementation, we use CWD Loss (channel-wise distillation loss)~\cite{shu2021channel} as the alignment loss, which shows better results than other loss functions. CWD Loss can be summarized as follows:
\begin{equation}
 \phi{\left(x_{c}\right) } =  \frac{\textup{exp}{(\frac{x_{c,i}}{\mathcal{T}} )}} {\sum_{i = 1}^{W\cdot H} \textup{exp}{(\frac{x_{c,i}}{\mathcal{T}} )} }~,
 \end{equation}
\vspace{-4pt}
 \begin{equation}
 \mathcal{L}_{\text{cwd}} = \frac{\mathcal{T}^2}{C}\sum_{c = 1}^{C}\sum_{i=1}^{H\cdot W}
	\phi (x_{T}^{c,i}) \cdot \log \Bigl[
	\frac{\phi(x_{T}^{c,i})}{\phi(x_{S}^{c,i})}
	\Bigr],
	\label{eq:cw-phi}
\end{equation}
where $c = 1,2,..., C$ indexes the channel, and $ i = 1,2,..., H\cdot W$ denotes the spatial location, $x^{T}$ and $x^{S}$ are the feature maps of the transformer branch and CNN branch, respectively. $\phi$ converts the feature activation into a channel-wise probability distribution, removing the influences of scales between the transformer and the compact CNN. To minimize $ \mathcal{L}_{\text{cwd}}$, $\phi(x_{S}^{c,i})$ should be large when $\phi(x_{T}^{c,i})$ is large. But when $\phi(x_{T}^{c,i})$ is small, the value of $\phi(x_{S}^{c,i})$ does not matter. This force the CNN to learn the distribution of the foreground salience, which contains the semantic information. $\mathcal{T}$ denotes a hyper-parameter called temperature. And the larger $\mathcal{T}$ is, the softer the probability distribution is. 

For more discussions and detailed ablation studies on the choices of alignment loss, please refer to appendix~\ref{section:B}.


\section{Experiments}

\subsection{Datasets and Implementation Details}
We conduct experiments of the SCTNet on three datasets, i.e.,~Cityscapes~\cite{cordts2016cityscapes}, ADE20K~\cite{zhou2017scene}, and COCO-Stuff-10K~\cite{caesar2018coco} to demonstrate the effectiveness of our method. For a fair comparison, we build our base model SCTNet-B with a comparable size to RTFormer-B/DDRNet-23/STDC2. Furthermore, we also introduce a smaller variant called SCTNet-S. We first pre-train our CNN backbones on ImageNet~\cite{deng2009imagenet}, then fine-tune it on semantic segmentation datasets. 
The semantic transformer branch in the training phrase can be any hierarchical transformer network.
In our implementation, we choose SegFormer as the transformer branch for all experiments. 
We measure the inference speed of all methods on a single NVIDIA RTX 3090.
All reported FPS results are obtained under the same input resolution for fair performance comparison unless specified.
For Cityscapes, we measure the speed implemented with both torch and tensor-RT. More details on model instantiation, metrics, and training settings of specific datasets can be found in appendix~\ref{section:A}. 

\subsection{Comparison with State-of-the-art Methods}

\noindent
{\bf Results on Cityscapes.} The corresponding results on Cityscapes\cite{cordts2016cityscapes} are shown in Table~\ref{tab:sample-tableSOTA}. Our SCTNet outperforms other real-time methods by a large margin and attains the best speed-accuracy trade-off with both tensorRT and Torch implementations. For example, our SCTNet-B-Seg100 achieves $80.5\%$ mIoU at $62.8$ FPS, which is a new state-of-the-art performance for real-time segmentation. Our SCTNet-B-Seg75 reaches $79.8\%$ mIoU, which is better than the state-of-the-art transformer-based bilateral network RTFormer-B and cnn-based bilateral network DDRNet-23 in accuracy but has a two times faster speed. Our SCTNet-B is faster at all input resolutions with better mIoU results than all other methods. Besides, our SCTNet-S also achieves a better trade-off compared with STDC2~\cite{fan2021rethinking}, RTFormer-S~\cite{wang2022rtformer}, SeaFormer-B~\cite{wan2023seaformer} and TopFormer-B~\cite{zhang2022topformer}. The origin speed with specific devices of more methods and the comparison of FPS on NVIDIA RTX 2080Ti GPU can be found in appendix~\ref{section:C}.

\begin{table}[t]
  \centering
  \begin{tabular}{l|c|c|c}
    \Xhline{1pt}
    Method & \#Params$\downarrow$ &FPS$\uparrow$ &mIoU(\%)$\uparrow$\\
    \hline
    FCN(MV2)     &9.8M &64.4$^*$   &19.7\\
    PSPNet(MV2) &13.7M &57.7$^*$ &29.6 \\
    DeepLabV3+(MV2)  &15.4M &43.1$^*$  &34.0 \\
    SegFormerB0  &3.8M    &84.4  &37.4 \\
    TopFormer-B &5.1M &96.2    &39.2\\
    SeaFormer-B &8.6M &44.5   &41.0 \\
    SegNext-T &4.3M  &60.3   &41.1 \\
    AFFormer-B &3.0M  &49.6   &41.8 \\
    RTFormer-S  &4.8M &95.2  &36.7\\
    RTFormer-B &16.8M &93.4 &42.1 \\
    \hline
    \textbf{SCTNet-S} &4.7M &\textbf{158.4}  &37.7 \\
    \textbf{SCTNet-B} &17.4M &145.1  &\textbf{43.0}  \\
    \Xhline{1pt}
  \end{tabular}
  \caption{\textbf{Comparisons with other state-of-the-art real-time methods on ADE20K.} The FPS is measured at resolution $512\times512$. * means speed from other papers, MV2 stands for MobileNetV2.} 
  \label{sample-tableADE20K}
  \vspace{-10pt}
\end{table}

\noindent
{\bf Results on ADE20K.} On ADE20K\cite{zhou2017scene}, our SCTNet achieves the best accuracy with the fastest speed. For instance, our SCTNet-B achieves $43.0\%$ mIoU at superior $145.1$ FPS, which is about 1.6 times faster than RTFormer-B~\cite{wang2022rtformer} with 0.9\% higher mIoU performance. Our SCTNet-S reaches $37.7\%$ mIoU while keeping the highest FPS among all other methods on ADE20K~\cite{zhou2017scene}. Considering the large variety of images and various semantic categories in ADE20K~\cite{zhou2017scene}, this outstanding results further also demonstrate the generalization capability of our SCTNet.

\noindent
{\bf Results on COCO-Stuff-10K.} The corresponding results on COCO-Stuff-10K are shown in Table~\ref{sample-COCO-Stuff}. SCTNet shows SOTA performance and maintains the highest inference speed on COCO-Stuff-10K in real-time semantic segmentation methods. With the input size $640 \times 640$, SCTNet-B achieves  $35.9\%$ mIoU at $141.5$ FPS, which is $0.6\%$ higher than RTFormer-B, and about 1.6 times faster. The comprehensive accuracy trade-off comparison on ADE20K and COCO-Stuff-10K can be found in Figure~\ref{suppfig:ADE_trade-off} and ~\ref{suppfig:COCO_trade-off} (Appendix \ref{app:sec:ade-coco-tradeoff}) respectively.

\begin{table}[t]

  \centering
  \begin{tabular}{l|c|c|c}
    \Xhline{1pt}
    Method & \#Params$\downarrow$ &FPS$\uparrow$ &mIoU(\%)$\uparrow$\\
    \hline
    PSPNet50    &-   &6.6$^*$       &32.6\\
    ICNet       &-  &35.7$^*$ &29.1 \\
    BiSeNetV2-L &-   &65.1       &28.7 \\
    TopFormer-B   &5.1M  &94.7       &33.4 \\
    SeaFormer-B &8.6M   &41.9       &34.1 \\
    AFFormer-B &3.0M  &46.5   &35.1 \\
    DDRNet23    &20.1M  &108.8     &32.1\\
    RTFormer-B  &16.8M  &90.9       &35.3 \\
    \hline
    \textbf{SCTNet-B} &17.4M &\textbf{141.5}  &\textbf{35.9}  \\
    \Xhline{1pt}
  \end{tabular}
  \caption{\textbf{Comparisons with other state-of-the-art real-time methods on COCO-Stuff-10K test set.} The FPS is measured at resolution $640\times640$. }
  \label{sample-COCO-Stuff}
  \vspace{-5pt}
\end{table}

\begin{figure*}[!t]
  \centering
  \setlength{\fboxsep}{0pt}
  \begin{tabular}{ccccc}
  \includegraphics[width=0.19\linewidth]{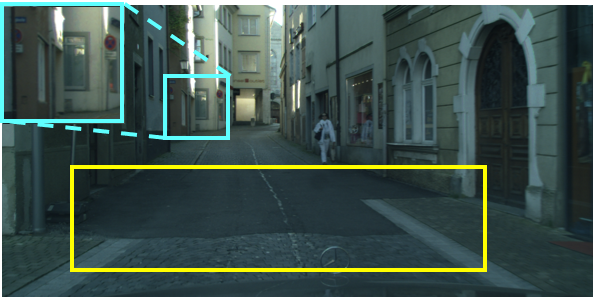}\label{1a} &
  \hspace{-12pt}
  \includegraphics[width=0.19\linewidth]{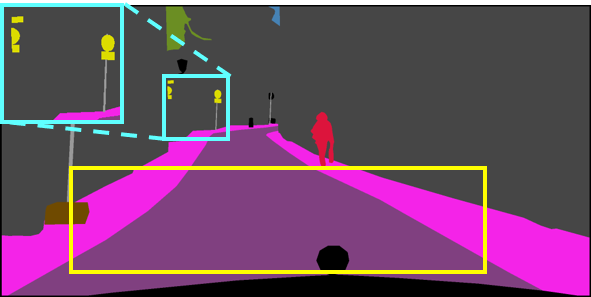}\label{1b} &
  \hspace{-12pt}
  \includegraphics[width=0.19\linewidth]{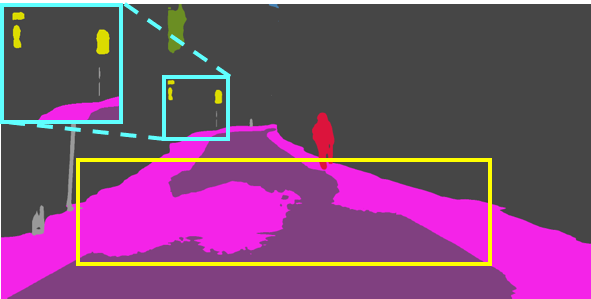}\label{1c} &
  \hspace{-12pt}
  \includegraphics[width=0.19\linewidth]{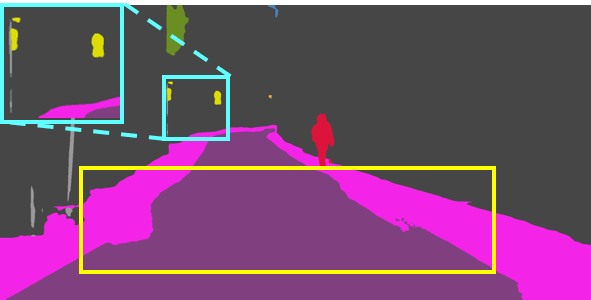}\label{1d} &
  \hspace{-12pt}
  \includegraphics[width=0.19\linewidth]{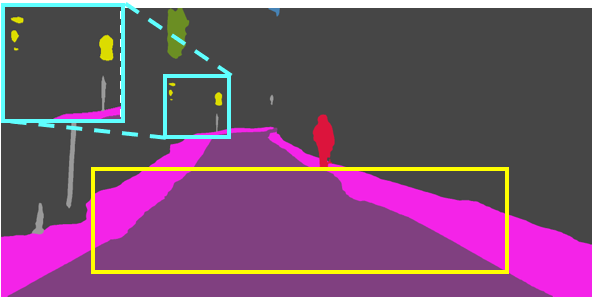}\label{1e} \\
  
  \includegraphics[width=0.19\linewidth]{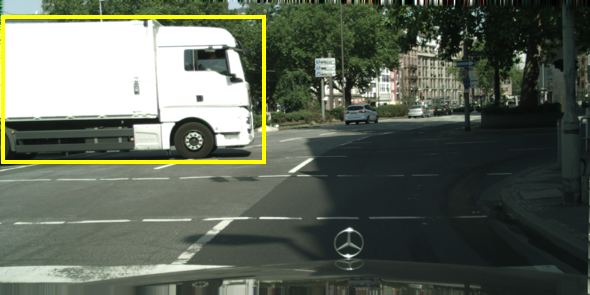}\label{2a} &
  \hspace{-12pt}
  \includegraphics[width=0.19\linewidth]{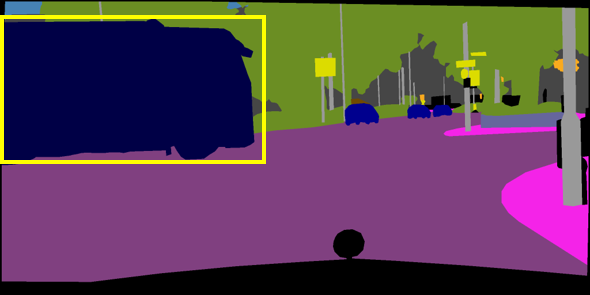}\label{2b} &
  \hspace{-12pt}
  \includegraphics[width=0.19\linewidth]{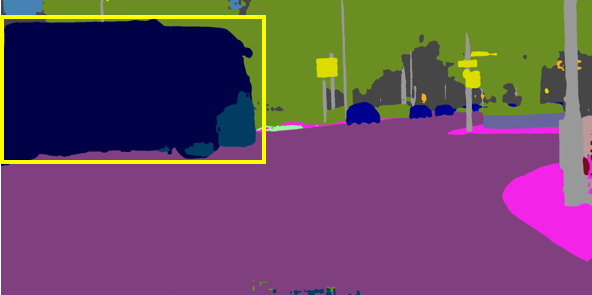}\label{2c} &
  \hspace{-12pt}
  \includegraphics[width=0.19\linewidth]{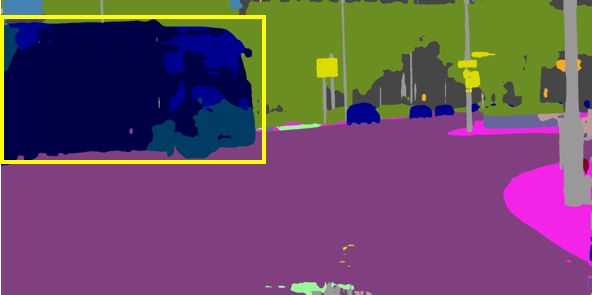}\label{2d} &
  \hspace{-12pt}
  \includegraphics[width=0.19\linewidth]{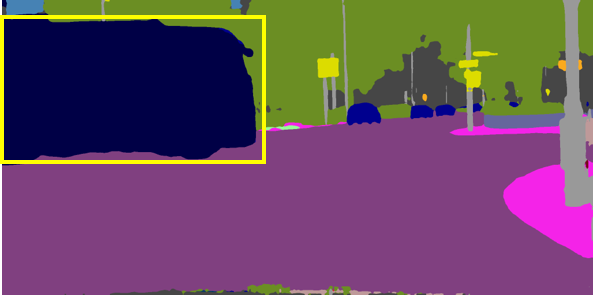}\label{2e} \\
  
  \includegraphics[width=0.19\linewidth]{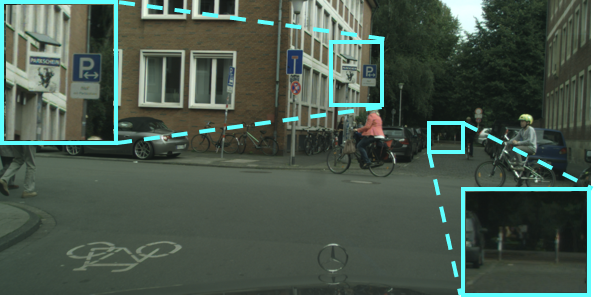}\label{3a} &
  \hspace{-12pt}
  \includegraphics[width=0.19\linewidth]{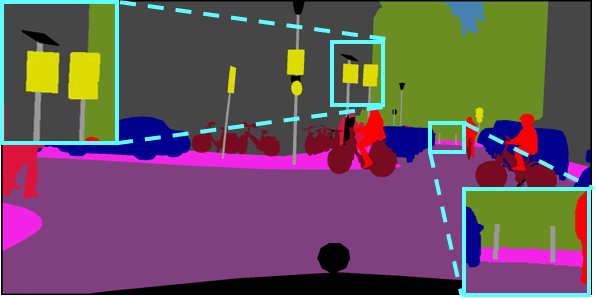}\label{3b} &
  \hspace{-12pt}
  \includegraphics[width=0.19\linewidth]{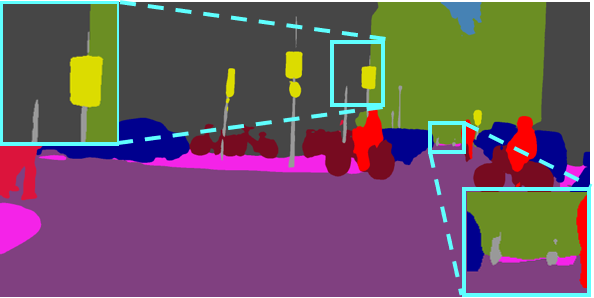}\label{3c} &
  \hspace{-12pt}
  \includegraphics[width=0.19\linewidth]{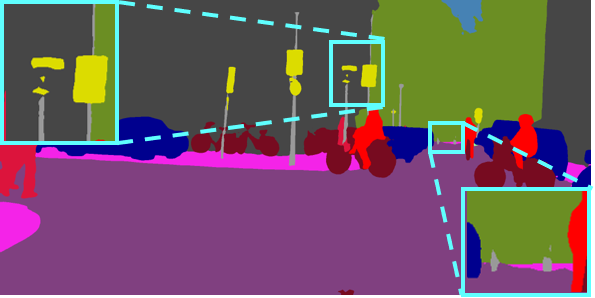}\label{3d} &
  \hspace{-12pt}
  \includegraphics[width=0.19\linewidth]{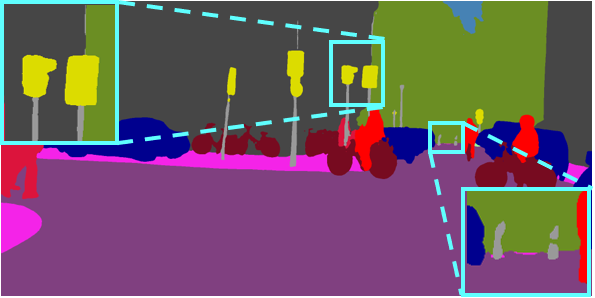}\label{3e} \\
  (a) Image & (b) GT & (c) DDRNet-23 & (d) RTFormer-B& (e) SCTNet-B\\
  \end{tabular}
  \caption{\textbf{Visualization results on Cityscapes validation set.} Compared with DDRNet-23\cite{pan2022deep} and RTFormer-B~\cite{wang2022rtformer}, SCTNet-B generates masks with finer details as highlighted in the light blue box and more accurate large-area predictions, as highlighted in the yellow box.
      }
  \label{fig:vis_cityscapes}
  
  \end{figure*}
  
\subsection{Ablation Study}

\noindent
{\bf Comparison on Different Types of Blocks.} To verify the effectiveness of our proposed CFBlock, we replace the CFBlocks with other kinds of convolution blocks and transformer blocks in real-time segmentation. For quick evaluations, all these results in Table~\ref{tab:Block Ablation} are not pre-trained on ImageNet. We select four kinds of blocks for comparison. As shown in Table~\ref{tab:Block Ablation}, our CFBlock outperforms the typical ResBlock and the lightweight SegFormer Block by a significant mIoU margin. Moreover, compared with the state-of-the-art GFABlock~\cite{wang2022rtformer} and MSCANBlock from SegNext~\cite{guo2022segnext}, our CFBlock get better speed and accuracy trade-off. Our CFBlock has 0.9\% higher mIoU than GFABlock and maintains the similar performance with fewer parameters and faster speed than MSCANBlock. This also demonstrates that our SCTNet can better mitigate the gap of semantic information between CNN and transformer while getting rid of the high computation cost.
\begin{table}
	\begin{center}
        \setlength{\tabcolsep}{1mm}
        \centering
        \begin{tabular}{l|c|c|c}
        \Xhline{1pt}
            Block
            &FPS$\uparrow$
            &mIoU(\%)$\uparrow$
            &param\\
            \hline
            ResBlock &66.7 &77.9 &15.3M\\
            SegFormerBlock &57.3 &77.7 &22.2M\\
            GFABlock &66.2 &78.5 &16.3M\\
            MSCANBlock &60.5    &79.3 &19.8M    \\
            \textbf{CFBlock (Ours)} &62.8 &\textbf{79.4} &17.4M\\
        \Xhline{1pt}
        \end{tabular}
      \end{center}
      \caption{\textbf{Comparison of different blocks.}}
         \label{tab:Block Ablation}
\vspace{-10pt}
\end{table}

\noindent
{\bf Effectiveness of the Semantic Information Alignment Module.}
Although our SIAM(semantic information alignment module) is closely related to the elaborately designed SCTNet, it can also improve the performance of other CNN and transformer segmentation methods. As presented in Table~\ref{tab:SIAM_Ablation}, employing our SIAM attains consistent improvements on SegFormer, SegNext, SeaFormer, and DDRNet, which proves the effectiveness and generalization capability of our proposed SIAM. At the same time, as representatives of the bilateral-branch transformer and the bilateral-branch CNN network, the improvements of SeaFormer and DDRNet are relatively slim. This may be attributed to the fact that their bilateral-branch network structure already benefits from the additional semantic branch. And this also confirms that the cooperation of our SIMA and training-only transformer does play the role of the semantic branch in the bilateral-branch network, leading to improvements in the accuracy of the single-branch network.
\begin{table}
\begin{center}
    \centering
    \resizebox{1.0\linewidth}{!}{
        \begin{tabular}{l|c|c|c}
        \Xhline{1pt}
            Block& Seg100(\%) &Seg75(\%) &Seg50(\%)\\
            \hline
            SegNext-T &79.8 &78.0 &-\\
            SegNext-T+SIAM &80.1(+0.3) &78.2(+0.2) &-\\
            \hline
            SegFormer-B0 &74.7 &74.4 &70.7\\
            SegFormer-B0+SIAM &77.3(+2.6) &76.8(+2.4) &72.5(+1.8)\\
            \hline
            SeaFormer-B &77.7 &- &72.2\\
            SeaFormer-B+SIAM &78.1(+0.4) &- &72.5(+0.3)\\
            \hline
            DDRNet-23 &79.5 &- &-\\
            DDRNet-23+SIAM &79.6(+0.1) &- &-\\
            \hline
            \textbf{SCTNet-B-SIAM} &78.5 &77.5 &75.2\\
            \textbf{SCTNet-B (Ours)} &80.5(+2.0) &79.8(+2.3) &76.5(+1.3)\\
        \Xhline{1pt}
        \end{tabular}
    }
\end{center}
    \caption{\textbf{Comparison of the effect of the SIAM.} 
    }
\label{tab:SIAM_Ablation}
\vspace{-5pt}
\end{table}

\noindent
{\bf Components Ablation.}
We explore the effect of the proposed components in Table~\ref{albation-Components}. Take Seg100 as an example, simply replacing the Resblock with our CFBlock brings a 2.1\% improvement of mIoU with little speed loss. The BFA leads to a 1.2\% higher mIoU, and the SDHA further attains a 0.8\% improvement of mIoU without sacrificing speed.

\begin{table}
  \centering
  \resizebox{1.0\linewidth}{!}{
  \begin{tabular}{l|c|c|c|c}
    \Xhline{1pt}
    Components &Seg100(\%)  &Seg75(\%) &Seg50(\%) &FPS(Seg100)\\
    \hline
    Baseline    &76.4       &76.0       &73.0       &66.7\\
    +CFBlock    &78.5(+2.1)  &77.5(+1.5)    &75.2(+2.2)   &62.8\\
    +BFA$^*$      &79.7(+1.2)   &79.1(+1.6)    &75.7(+0.5)   &62.8\\
    +SDHA       &80.5(+0.8)   &79.8(+0.7)       &76.5(+0.8) &62.8\\
    \Xhline{1pt}
  \end{tabular}
  }
  
  \caption{\textbf{Ablation studies on the components of SCTNet}} 
  \label{albation-Components}
\vspace{-10pt}
\end{table}

\noindent
{\bf More Ablation Studies.} For more detailed ablation studies on the design of CFBlock, the discussion of semantic alignment loss, and the choice of semantic transformer branch in training phrase, please refer to the appendix~\ref{section:B}.

\subsection{Visualization Results}
Figure~\ref{fig:vis_cityscapes} shows visualization results on Cityscapes~\cite{cordts2016cityscapes} validation set. Compared with DDRNet and RTFormer, our SCTNet provides not only better results for those classes with large areas like roads, sidewalks, and big trucks but also more accurate boundaries for small or thin objects such as poles, traffic lights, traffic signs, and cars. This indicates that SCTNet extracts high-quality long-range context while preserving fine details. More visualization results on Cityscapes and ADE20K can be found in appendix~\ref{section:E}.

\section{Conclusion}
In this paper, we propose SCTNet, a novel single-branch architecture that can extract high-quality long-range context without extra inference computation cost. Extensive experiments demonstrate that SCTNet achieves new state-of-the-art results. Moreover, by demonstrating the efficiency of SCTNet, we provide a novel insight for the semantic branch in the bilateral-branch network and a new way to boost the real-time segmentation community by not only adopting the structure of the transformer but also unitizing its knowledge.

\section{Acknowledgments}
This work is supported by Hubei Provincial Natural Science Foundation of China No.2022CFA055 and the National Natural Science Foundation of China No.62176097.

\bibliography{aaai24}

\newpage
\appendix


\section*{Appendix} 
The structure of this supplementary material can be summarized as follows. Section~\ref{section:A} provides more implementation details. Section~\ref{section:B} shows more detailed ablation studies on the design of CFBlock, semantic alignment loss, and semantic transformer branch in training phrase. Section~\ref{section:C} presents the comparison on NVIDIA RTX 2080Ti and original speed with specific devices of more methods. Section~\ref{section:D} gives some discussions about the proposed SCTNet. Moreover, more visualization results are displayed in Section~\ref{section:E}.

\section{More Implementation Details}\label{section:A}
\subsection{ImageNet Pre-training}
For a fair comparison, the CNN backbone of the proposed SCTNet is pre-trained on ImageNet-1K~\cite{deng2009imagenet} using 8 GPUs. In the pre-training phase, we build our code based on the MMClassification, following the training settings of swin-transformer~\cite{liu2021swin} on ImageNet-1K. The main details of the pre-training settings are presented in Table~\ref{supp-tab:settings}.

\begin{table}[!h]
\begin{center}
\caption{\textbf{Training settings on ImageNet-1K.}}
\begin{tabular}{l|l}
\toprule
config & value \\
\hline
optimizer & AdamW \\
base learning rate & 0.001\\
weight decay & 0.05\\
optimizer momentum & $\beta_1, \beta_2{=}0.9, 0.999$ \\
learning rate schedule & cosine decay \\
minimum learning rate & 1e-5 \\
warmup epochs & 20 \\
warmup learning rate & 1e-6 \\
training epochs & 300  \\
batch size & 1024 \\
augmentation & RandAug(9, 0.5) \\
random resized crop & 224 \\
random flip & 0.5\\
mixup & 0.8 \\
cutmix & 1.0 \\
random erasing & 0.25 \\

\bottomrule
\end{tabular}
\vspace{-10pt}
\label{supp-tab:settings}
\end{center}
\end{table}

\subsection{Datasets and Implementation Details}
{\bf Cityscapes.} Cityscapes~\cite{cordts2016cityscapes} is an urban street semantic scene parsing dataset from a car's perspective. It contains 5,000 fine annotated images with 19 categories for segmentation, which are split into training, validation, and test sets, with 2,975, 500, and 1,525 images, respectively. For Cityscapes~\cite{cordts2016cityscapes}, we train all models using AdamW~\cite{loshchilov2017decoupled} optimizer with the initial learning rate of 0.0004 and the weight decay of 0.0125. We apply the poly learning policy with the power of 0.9 to drop the learning rate and implement the data augmentation method, including random cropping, random scaling, and random horizontal flipping. For Seg75\& Seg50, we adopt random cropping into $1024 \times 512$ and random scaling in the range of 0.25 to 1.5, while $1024 \times 1024$ and from 0.5 to 2.0 for Seg100. All models are trained with 160k iterations, a batch size of 16. For the Tensor-RT speed, we apply TensorRT v8.2.3 for acceleration.

\noindent
{\bf ADE20K.} ADE20K~\cite{zhou2017scene} is a scene parsing dataset with 150 fine-grained semantic classes, which is split into 20K, 2K, and 3K images for training, validation, and testing, respectively. For ADE20K, all models are trained with a batch size of 32 for 160k iterations. Except for the cropping size of $512 \times 512$ and the initial learning rate of 0.0005, all other training settings are identical to the training settings of Seg100 for Cityscapes~\cite{cordts2016cityscapes}.

\noindent
{\bf COCO-Stuff.} COCO-Stuff~\cite{caesar2018coco} . COCO-Stuff contains 171 semantic classes and 10K(9K for training and 1K for testing) difficult samples with pixel-level stuff annotations that are collected in COCO. We use AdamW optimizer with weight decay of 0.00006 and an initial learning rate of 0.01. Data augmentation includes random cropping into $640 \times 640$ and random scaling in the range of 0.5 to 2.0. All other training details are the same as the training settings of Seg100 for Cityscapes~\cite{cordts2016cityscapes}.

\noindent
{\bf Metric.} We adopt mIoU~(mean intersection over union), FPS~(Frames Per Second),  and parameters as the evaluation metrics. mIoU is a widely used evaluation metric in semantic segmentation tasks. It quantifies the degree of overlap between the predicted segmentation and the ground truth segmentation by computing the ratio of the intersection to the union of the two sets of pixels. The mIoU score is obtained by averaging the IoU scores across all classes in the dataset. It provides a measure of the overall accuracy of the segmentation model. Frames Per Second (FPS) is a measure of the number of frames that the network can process per second, which is affected by the specific device. FPS can intuitively measure the inference speed of the network. Parameters refers to the total number of parameters that must be trained during model training. This metric is commonly used to measure the size of a model. On resource-constrained edge devices, the parameter volume is a critical factor to consider. FLOPs is not chosen as one of the evaluation metrics because it can not directly estimate the inference speed.

\begin{table}
	\centering
        \caption{\textbf{Ablation studies on the kernel size of the learnable kernels in CFBlock}.
        }
        \setlength{\tabcolsep}{1mm}
        \centering
        \begin{tabular}{l|c|c}
        \Xhline{1pt}
            Kernel Size
            &FPS$\uparrow$
            &mIoU(\%)$\uparrow$ \\
            \hline
            $1 \times 1$ &67.4 &78.4\\
            $1 \times 1 + 1 \times 1$ &65.9 &78.7\\
            $1 \times 3 + 3 \times 1$ &64.8 &79.0\\
            $1 \times 5 + 5 \times 1$ &63.3 &79.1\\
            $1 \times 7 + 7 \times 1$ &62.8 &\textbf{79.4}\\
            $1 \times 9 + 9 \times 1$ &61.9 &79.3\\
            $7 \times 7$ &59.8 &78.8\\
        \Xhline{1pt}
        \end{tabular}
		\label{supptab:Kernel Size Ablation}
\end{table}

\begin{table*}[!t]
	\setlength{\fboxrule}{0pt}
	\caption{\textbf{Ablations on the semantic alignment loss, including types, locations and weight.}}
	\label{last-three}
	\vspace{-10pt}
	\begin{center}

\begin{subtable}[t]{0.28\textwidth}
	\begin{center}
        \caption{Ablation on different types of loss. }
        \vspace{1pt}
        \setlength{\tabcolsep}{1mm}
        \centering
        \resizebox{0.97\linewidth}{!}{
        \begin{tabular}{l|ccc}
        \toprule
            Loss Type
            &Seg100
            &Seg75
            &Seg50 \\
            \hline
            Baseline     &78.5 &77.5 &75.2\\
            L2 loss    &77.8 &76.8 &73.0\\
            KL loss  &78.6 &77.6 &75.2\\
            MI loss  &\textbf{79.7}     &79.0     & \textbf{75.7}     \\
            CWD loss &\textbf{79.7} &\textbf{79.1} &\textbf{75.7}\\
        \bottomrule
        \end{tabular}
        }
		\label{tab:loss type}
	\end{center}
\end{subtable}
\hfill
\hspace{3pt}
\begin{subtable}[t]{0.35\textwidth}
	\begin{center}
        \caption{Ablation on the location of alignment loss. }
        \setlength{\tabcolsep}{1mm}
        \centering
        \resizebox{0.95\linewidth}{!}{
        \begin{tabular}{cccc|c}
        \toprule
            logits
            &decoder
            &stage4
            &stage3 
            &mIoU(\%)$\uparrow$\\
            \hline
                        &       &       &   &78.2\\
            \checkmark  &       &       &   &78.8(+0.6)\\
            \checkmark  &\checkmark   &  &      &79.1(+0.9)\\
            \checkmark  &\checkmark  &\checkmark   &  &79.3(+1.1)\\
            \checkmark  &\checkmark  &\checkmark    &\checkmark  &\textbf{79.7(+1.5)}\\
        \bottomrule
        \end{tabular}
        }
		\label{tab:loss location}
	\end{center}
\end{subtable}
\hfill
\begin{subtable}[t]{0.35\textwidth}
	\begin{center}
        \caption{Ablation on the weight of alignment loss. }
        \vspace{1pt}
        \setlength{\tabcolsep}{1mm}
        \centering
        \resizebox{0.82\linewidth}{!}{
        \begin{tabular}{l|ccc}
        \toprule
            Loss Weight
            &Seg100
            &Seg75
            &Seg50 \\
            \hline
            0,0,0,0     &78.5 &77.5 &75.2\\
            3,0,0,0     &78.8 &78.5 &75.3\\
            3,10,10,10  &79.5 &78.9 &75.5\\
            3,15,15,15  &\textbf{79.7} &\textbf{79.1} &\textbf{75.7}\\
            3,20,20,20  &79.1 &78.6 &75.0\\
        \bottomrule
        \end{tabular}
        }
		\label{tab:loss weight}
	\end{center}
\end{subtable}

\end{center}
\end{table*}

\subsection{Model Instantiation}
We build our base model SCTNet-B with a comparable size to that of RTFormer-B/DDRNet-23/STDC2. Furthermore, we also introduce a smaller variant called SCTNet-S. The network architecture hyper-parameters of the two model variants are as follows:
\begin{itemize}
    \item SCTNet-B: $C=\{64,128,256,512\}$, layer numbers $=\{2,2,3,2\}$, $N = 64$, $k=7$.
    \item SCTNet-S: $C=\{32,64,128,256\}$, layer numbers $=\{2,2,3,2\}$, $N = 64$, $k=7$.
\end{itemize}
where $C$ denotes the channel number of each stage, $N$ is the number of key/value in Conv-Former blocks, and $k$ denotes the kernel size of learnable kernels in Conv-Former blocks.

 The two model variants utilize the same loss setting. We use CE loss~\cite{milletari2016v} and CWD loss~\cite{shu2021channel} as our overall training loss, which can be formulated as follows: 
\begin{equation}\label{Eq:TrainingLoss}
\begin{aligned}
\mathcal{L}_{\text{all}} = \lambda_{\text{main}}\mathcal{L}_{\text{ce}} + \lambda_{\text{aux}}\mathcal{L}_{\text{ce}} + \sum_{i=0}^{3}\lambda_{\text{cwd}}^{i}\mathcal{L}_{\text{cwd}}^{i}  
\end{aligned}
\end{equation}
where $\lambda_{\text{main}}$ and $\lambda_{\text{aux}}$ denote the loss weights of the decoder head and auxiliary head in stage 2, respectively. $\lambda_{\text{cwd}}$ denotes the loss weight of the context guidance head. We set the hyper-parameters to $\lambda_{\text{main}} = 1.0$, $\lambda_{\text{aux}} = 0.4$, and $\lambda_{\text{cwd}} = [3,15,15,15]$ for output logits, features of decoder, stage 4 and stage 3, respectively. Follow~\cite{shu2021channel}, we set temperature $\mathcal{T} = 4$ .

As for the semantic transformer branch in the training phrase, it can be any hierarchical transformer. In our implementation, we choose SegFormer~\cite{xie2021segformer} as the transformer teacher for all experiments. Specifically, we choose SegFormer-B3 for SCTNet-B and SegFormer-B2 for SCTNet-S.

\section{More Ablation Experiments}\label{section:B}
\subsection{Ablation Studies on the design of CFBlock}
To further clarify the structure of our Conv-Former Block, we conduct some ablation studies on the details of the proposed convolutional attention module. Taking efficiency into consideration, we implement the convolutional attention module with stripe convolution rather than standard convolutions. More specifically, we utilize a $1 \times 7$ and a $7 \times 1$ convolution to approximate a $7 \times 7$ convolution layer. Table~\ref{supptab:Kernel Size Ablation} shows the ablation studies on the kernel size of the learnable kernels. As shown in Table~\ref{supptab:Kernel Size Ablation}, when kernel size is set to 7, \ie using a $1 \times 7$ and a $7 \times 1$ stripe convolutions, the SCTNet reaches the highest mIoU, 1.0\% higher than a single $1 \times 1$ convolution, while the FPS only decreases by 4.6. When the kernel size is larger than 7, the mIoU stops to improve, but the computational cost still increases. Therefore, we set the hyper-parameter kernel size to 7 in all experiments. Note that although the $7 \times 7$ standard convolution owns the slowest speed, it gets a lower mIoU compared with a $1 \times 7$ and a $7 \times 1$ stripe convolutions. All experiments in Table~\ref{supptab:Kernel Size Ablation} are without pre-training.

\subsection{Ablation Studies on Semantic Alignment Loss}
We conduct extensive experiments on the types, locations, and weights of the alignment loss. 

The proposed semantic information alignment module aims at instructing the CNN to learn the long-range semantic information from the transformer. Therefore, the loss which can supervise the learning of context information is more suitable for our alignment loss. According to the results in Table~\ref{last-three}(a), CWD loss has positive effects on the performance, while L2 loss decreases the accuracy compared with the baseline. At the same time, KL loss shows almost the same performance before the alignment. This phenomenon could be attributed to the fact that CWD loss converts the feature activation into channel-wise probability distribution so that it focuses on the overall information alignment in each channel, which may contain rich global context. While L2 loss and KL loss directly align feature maps in an element-wise manner without regarding feature maps as an ensemble. Compared with KL loss, L2 loss adopts harder alignment, leading to worse performance. MI loss refers to mutual information loss, which is a measure of mutual dependence between two random variables. We use the technique of variational estimation in VID~\cite{ahn2019variational} to estimate the mutual information of the distribution of each channel and achieve similar performance to CWD loss. Both MI and CWD losses are applied to the channel-wise distribution. Therefore, what matters in our SCTNet is the alignment of the distribution between channels rather than the specific form of the similarity measurement. This is because SCTNet aims to align the long-range context of the semantic branch, which is contained in the overall channel distribution rather than in each pixel or local area. Considering that the form of CWD loss is simpler and easier for readers to understand, we ultimately adopted CWD loss. 

Table~\ref{last-three}(b) shows the influence of the alignment loss location. More long-range semantic information may exist in deep stages of hierarchy CNN, and the results also indicate that applying semantic alignment on stage3\&4, decoder, and output logits is the best setting for the mIoU performance. Note that the location of alignment loss does not make any difference in the inference FPS. 

At last, we explore the weight of alignment loss in Table~\ref{last-three}(c). For output logits, 3 is the best choice following~\cite{shu2021channel}. For features, we find that 15 is the most suitable. Smaller weights can not fully exploit the high-quality semantic information of the transformer, while larger weights compete with the CE loss. Moreover, we observe that when the weights are too large, such as 50, the value of training loss fluctuates dramatically, leading to low accuracy. Note that all of these ablation studies are conducted without the shared decoder head alignment.

\subsection{Ablation Studies on Semantic Branch in Training Phrase} 
We also conduct ablation experiments on different training-only semantic branches to screen the appropriate ones. Table~\ref{supptab:Semantic Branch Ablation} presents the influence of four types of semantic branches: OCRNet, InternImage, Swin, and SegFormer. We adopt these types of branches with similar accuracy, and the training settings are the same. The consistent improvement observed in the experiments demonstrates the effectiveness of the proposed SIAM. Among the candidate branches, SegFormer and InternImage showed the most promising improvement. However, SegFormer requires less training time and memory compared to others. Therefore, we selected SegFormer as the final semantic branch for the training phase.

\begin{table}
	\centering
        \caption{\textbf{Ablation Studies on Semantic Branch in Training Phrase}.
        }
        \setlength{\tabcolsep}{1mm}
        \centering
        \begin{tabular}{l|c|c|c}
        \Xhline{1pt}
            Semantic Branch
            &Seg100
            &Seg75
            &Seg50 \\
            \hline
            OCRNet     &80.0 &79.2 &76.4\\
            InternImage  &\textbf{80.5} &79.6 &\textbf{76.5}\\
            SwinTransformer  &79.8 &79.1 &76.0\\
            SegFormer &\textbf{80.5} &\textbf{79.8} &\textbf{76.5}\\
        \Xhline{1pt}
        \end{tabular}
		\label{supptab:Semantic Branch Ablation}
\end{table}

\section{Comparison with More Methods}\label{section:C}

\begin{figure}[tb]
\begin{center}
   \includegraphics[width=1.0\linewidth]{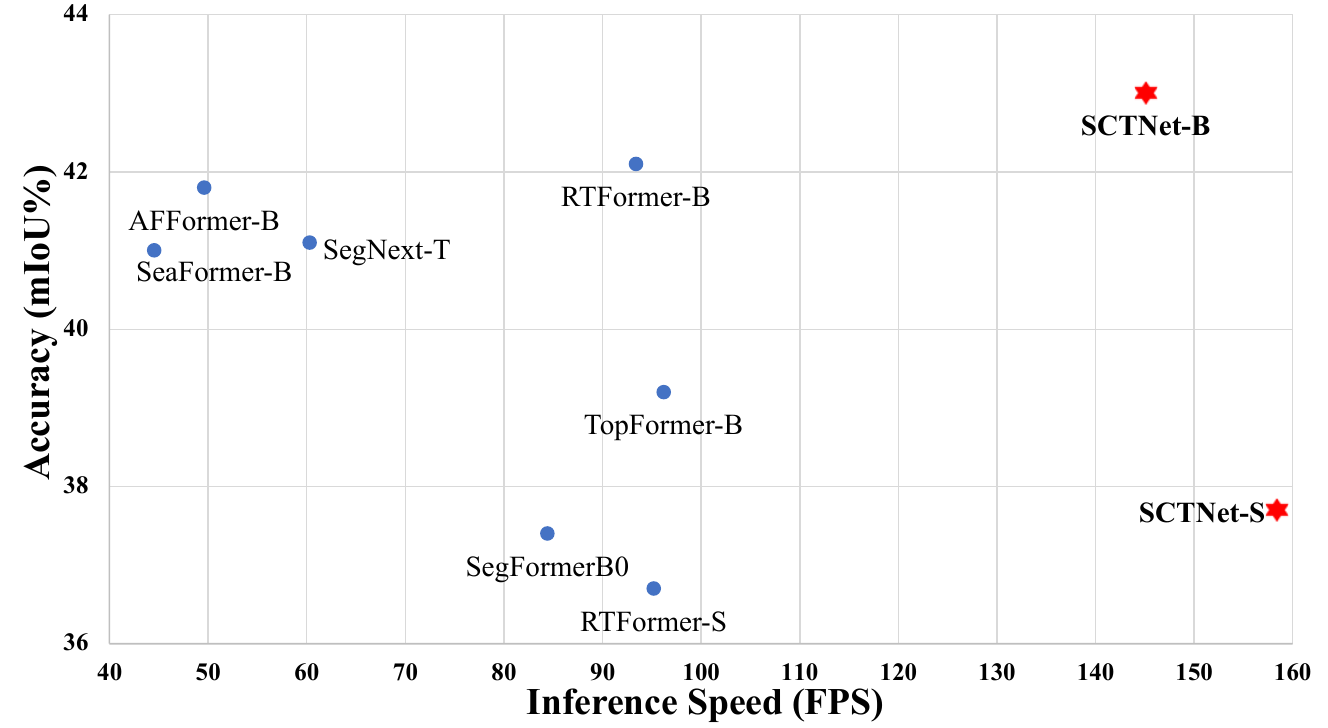}
\end{center}
   \caption{\textbf{The speed-accuracy performance on ADE20K validation set.} }
\label{suppfig:ADE_trade-off}
\vspace{-10pt}
\end{figure}

\begin{figure}[tb]
\begin{center}
   \includegraphics[width=1.0\linewidth]{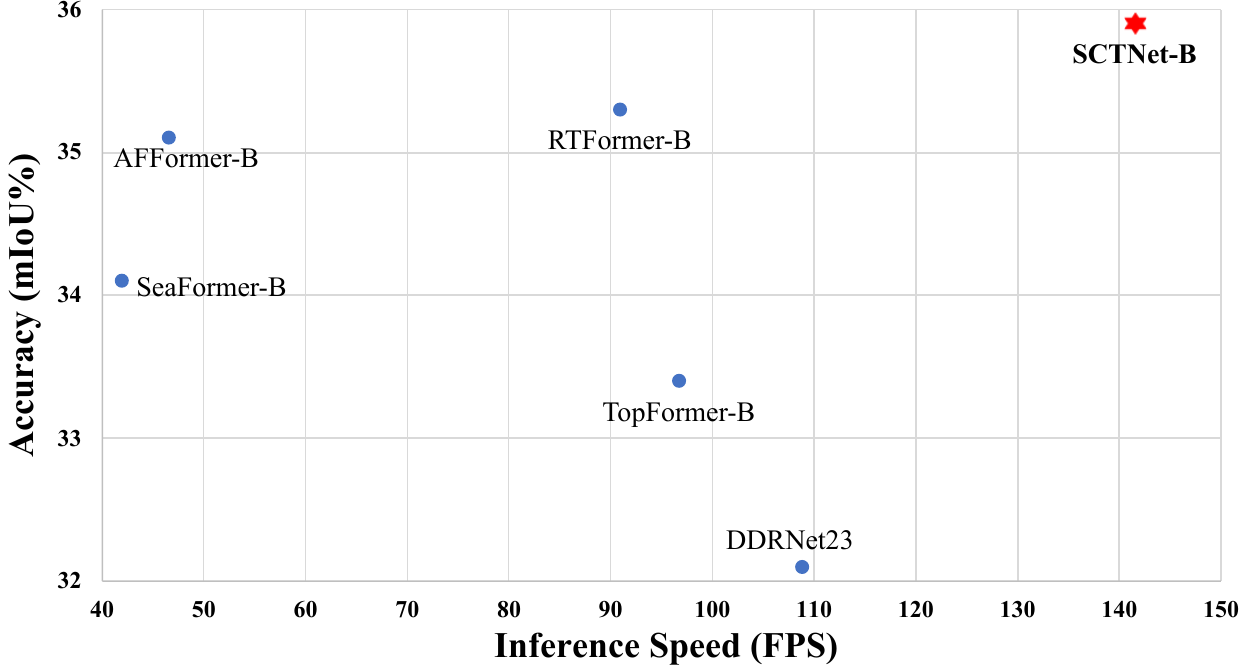}
\end{center}
   \caption{\textbf{The speed-accuracy performance on COCO-Stuff-10K test set.} }
\label{suppfig:COCO_trade-off}
\end{figure}

\subsection{Trade-off on ADE20K and COCO-Stuff-10K}\label{app:sec:ade-coco-tradeoff}

Most CNN-based real-time segmentation methods mainly focus on Cityscapes~\cite{cordts2016cityscapes}, while hardly taking ADE20K~\cite{zhou2017scene} and COCO-Stuff-10K~\cite{caesar2018coco} into consideration. These datasets contain numerous images with over 150 classes, making them challenging for lightweight CNN-based models. However, recently some transformer-based real-time segmentation methods have shown promising improvement on ADE20K and COCO-Stuff-10K. Although our SCTNet is a CNN-based network, it aligns the semantic information with the transformer in the training phase. Therefore, the SCTNet also shows competitive performance on ADE20K and COCO-Stuff-10K. As present in Figure~\ref{suppfig:ADE_trade-off} and Figure~\ref{suppfig:COCO_trade-off}, our SCTNet outperforms other transformer-based and CNN-based methods in real-time segmentation by a significant margin.

\begin{table*}[t]
  \caption{\textbf{Comparisons with bilateral real-time methods on Cityscapes val set.} All the speed is measured on a single RTX 2080Ti.}
  \label{tab:suppSOTA}
    \centering
  \begin{tabular}{c l|c|c|c|c|c}
    \Xhline{1pt}
    Method & Reference &\#Params$\downarrow$ &Resolution &FPS(TRT)$\uparrow$ &FPS(Torch)$\uparrow$ &mIoU(\%)$\uparrow$  \\
    \hline
    \multicolumn{2}{c|}{\textcolor{gray}{CNN-based Bilateral Networks}} & &  & & &\\
    BiSeNet-ResNet18  &ECCV~\citeyear{yu2018bisenet}  &49.0M  &$1536 \times 768$     &112.3 &72.4 &74.8  \\
    BiSeNetV2-L      &IJCV~\citeyear{yu2021bisenet}   &-      &$1024 \times 512$    &70.3 &35.2   &75.8   \\
    STDC1-Seg75 &CVPR~\citeyear{fan2021rethinking} &14.2M &$1536 \times 768$ &140.9 &72.2 &74.5 \\
    STDC2-Seg75 &CVPR~\citeyear{fan2021rethinking} &22.2M &$1536 \times 768$ &104.1 &57.2  &77.0 \\
    STDC1-Seg50 &CVPR~\citeyear{fan2021rethinking} &14.2M &$1024 \times 512$ &290.7 &143.6 &72.2 \\
    STDC2-Seg50 &CVPR~\citeyear{fan2021rethinking} &22.2M &$1024 \times 512$ &236.1 &89.8  &74.2 \\
    DDRNet-23-S &TIP~\citeyear{pan2022deep} &5.7M &$2048 \times 1024$ &93.2 &85.5 &77.8 \\
    DDRNet-23 &TIP~\citeyear{pan2022deep} &20.1M &$2048 \times 1024$ &70.2 &36.7 &\textbf{79.5} \\
    \hline
    \multicolumn{2}{c|}{\textcolor{gray}{Transformer-based Bilateral Networks}} & &  & & &\\
    TopFormer-B-Seg100 &CVPR~\citeyear{zhang2022topformer}    &5.1M &$2048 \times 1024$ &85.3 &64.2 &76.3 \\
    TopFormer-B-Seg50  &CVPR~\citeyear{zhang2022topformer}    &5.1M &$1024 \times 512$ &310.2 &90.8 &70.7\\
    SeaFormer-B-Seg100 &ICLR~\citeyear{wan2023seaformer}    &8.6M &$2048 \times 1024$ &77.6 &36.7 &77.7 \\
    SeaFormer-B-Seg50 &ICLR~\citeyear{wan2023seaformer} &8.6M &$1024 \times 512$ &223.5 &43.9 &72.2 \\
    RTFormer-S &NeurIPS~\citeyear{wang2022rtformer}    &4.8M &$2048 \times 1024$ &- &70.9 &76.3 \\
    RTFormer-B &NeurIPS~\citeyear{wang2022rtformer}    &16.8M &$2048 \times 1024$ &- &30.8 &\textbf{79.3} \\
    \hline
    \textbf{SCTNet-S-Seg50}  &Ours &4.6M &$1024 \times 512$ &327.5 &\textbf{225.1} &72.8\\
    \textbf{SCTNet-S-Seg75} &Ours &4.6M &$1536 \times 768$ &156.6 &147.9 &76.1 \\
    \textbf{SCTNet-B-Seg50}  &Ours &17.4M &$1024 \times 512$ &244.4 &136.9 &76.5 \\
    \textbf{SCTNet-B-Seg75}  &Ours &17.4M &$1536 \times 768$ &115.2 &65.7 &79.8\\
    \textbf{SCTNet-B-Seg100} &Ours &17.4M &$2048 \times 1024$ &76.7  &42.0 &\textbf{80.5} \\
    \Xhline{1pt}
  \end{tabular}
\end{table*}

\begin{figure}[tb]
\begin{center}
   \includegraphics[width=\linewidth]{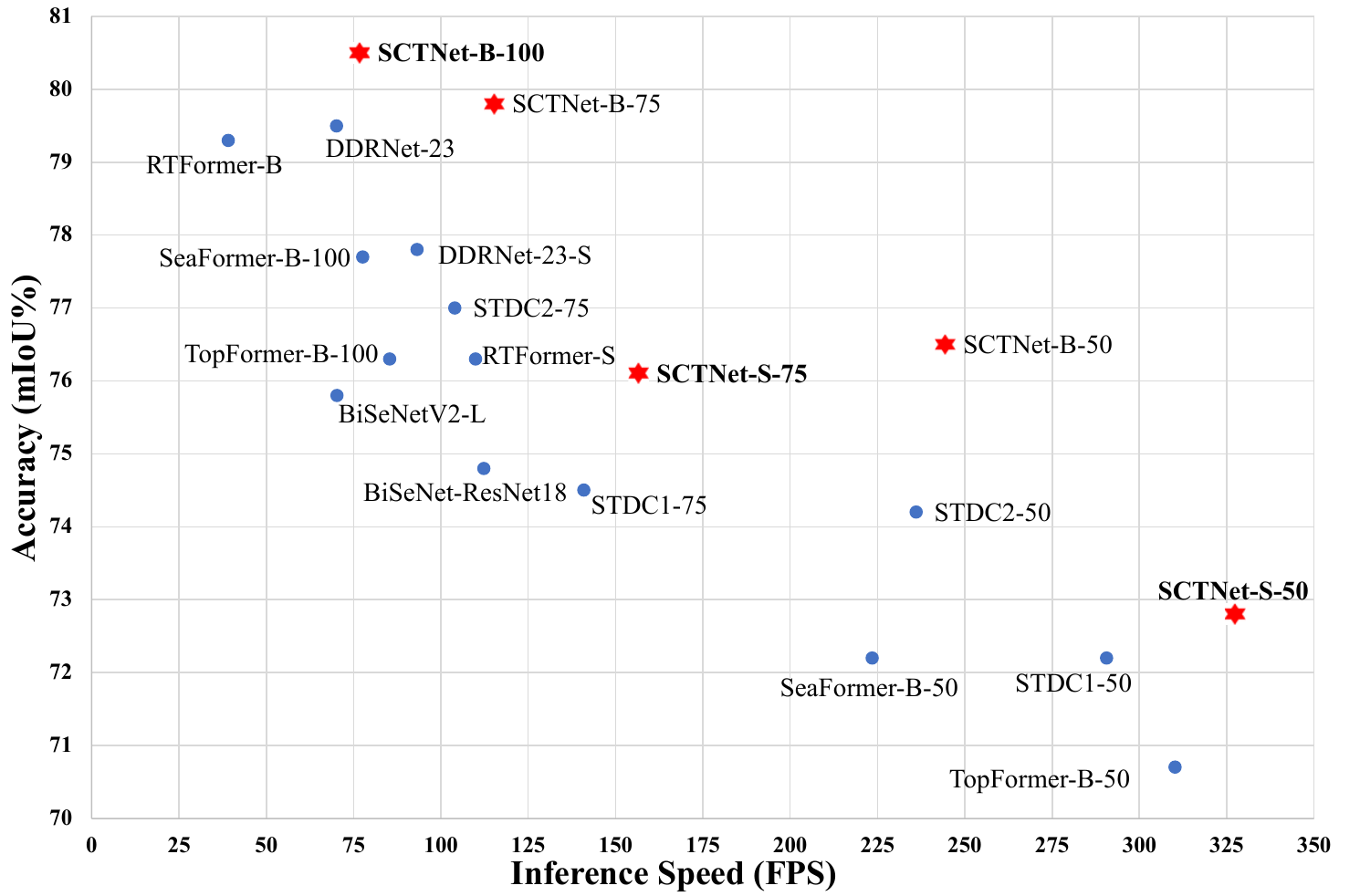}
\end{center}
   \caption{\textbf{Speed(TRT)-Accuracy performance on Cityscapes validation set using a single NVIDIA RTX 2080Ti}. Our methods are presented in red starts, while others are presented in blue dots. Our SCTNet establishes a new state-of-the-art speed-accuracy trade-off.}
\label{suppfig:figure1_trade-off}
\end{figure}

\subsection{Trade-off on RTX 2080Ti}\label{app:rtx2080}
 To further demonstrate the high efficiency of our SCTNet on different devices, we also measure the inference speed of our SCTNet and the mentioned bilateral methods using a single NVIDIA RTX 2080Ti. The results are presented in Table~\ref{tab:suppSOTA}. 
 To clearly show the efficiency of our proposed SCTNet, we also present the accuracy-speed comparison in Figure~\ref{suppfig:figure1_trade-off}. Using the same NVIDIA RTX 2080Ti, our SCTNet-B also outperforms existing bilateral real-time semantic segmentation networks by a large margin. At the same time, our SCTNet-S keeps a competitive trade-off of speed and accuracy with these methods.
 Note that, TensorRT does not support some operations used in the RTFormer~\cite{wang2022rtformer}, but the acceleration details of RTFormer are not available. So we report the speed from its original paper in Figure~\ref{suppfig:figure1_trade-off}, which is also deployed on RTX 2080Ti.

\subsection{More Methods with Specific Devices}
In the main paper, we measured all speeds on a single RTX 3090 for fair comparison. And because of space limitations, we only compare recent state-of-the-art works. Therefore, we perform a more comprehensive comparison in Table~\ref{tab:Sup_overall}. Note that all speeds mentioned here are obtained from the original paper with corresponding devices.

\begin{table*}[t]
  \caption{\textbf{ More comprehensive accuracy and speed performance comparison on Cityscapes val set.} All the speed in this table is from the original paper with corresponding devices. $^*$ means it uses some acceleration methods, not the original torch speed.
  }
  \label{tab:Sup_overall}
    \centering
  \begin{tabular}{c|l|c|c|c|c|c}
    \Xhline{1pt}
    Method & GPU &\#Params$\downarrow$ &Resolution &FPS(TRT)$\uparrow$ &FPS(Torch)$\uparrow$ &mIoU(\%)$\uparrow$ \\
    \hline
    ERFNet  &TitanX M   &20M &$2048 \times 1024$ &- &41.7 &70.0\\
    Fast-SCNN  &TitanXp   &1.1M  &$2048 \times 1024$ &- &123.5 &68.6\\
    CAS  &TitanXp   &-  &$1536 \times 768$ &- &108 &71.6\\
    GAS  &TitanXp   &-  &$1536 \times 768$ &- &108.4 &72.4\\
    FaPN-R34  &Titan RTX   &-  &$2048 \times 1024$ &- &30.2 &78.5\\
    AFFormer-B &V100 &3.0M &$2048 \times 1024$ &- &22 &78.7 \\
    \hline
    FasterSeg &GTX1080Ti  &4.4M &$2048 \times 1024$ &163.9 &- &73.1 \\
    SwiftNetRN-18 &GTX1080Ti  &11.8M &$2048 \times 1024$ &- &104.0 &75.5 \\
    PP-LiteSeg-T2  &GTX1080Ti     &-  &$1536 \times 768$  &143.6 &- &76.0\\
    PP-LiteSeg-B2  &GTX1080Ti     &-  &$1536 \times 768$  &102.6 &- &78.2\\
    DF2-Seg1  &GTX1080Ti  &-  &$1536 \times 768$     &67.2 &- &75.9  \\
    DF2-Seg2  &GTX1080Ti  &-  &$1536 \times 768$     &56.3 &- &76.9  \\
    BiSeNet-ResNet18  &GTX1080Ti  &49.0M  &$1536 \times 768$     &- &65.5 &74.8  \\
    BiSeNetV2-L      &GTX1080Ti   &-      &$1024 \times 512$    &47.3 &-    &75.8   \\
    STDC1-Seg75 &GTX1080Ti &14.2M &$1536 \times 768$ &126.7 &- &74.5 \\
    STDC2-Seg75 &GTX1080Ti &22.2M &$1536 \times 768$ &97.0 &-  &77.0 \\
    STDC1-Seg50 &GTX1080Ti &14.2M &$1024 \times 512$ &250.4 &- &72.2 \\
    STDC2-Seg50 &GTX1080Ti &22.2M &$1024 \times 512$ &188.6 &-  &74.2 \\
    \hline
    DDRNet-23-S &RTX 2080Ti  &5.7M &$2048 \times 1024$ &- &101.6 &77.8 \\
    DDRNet-23 &RTX 2080Ti  &20.1M &$2048 \times 1024$ &- &37.1 &79.5 \\
    RTFormer-S &RTX 2080Ti    &4.8M &$2048 \times 1024$ &- &110.0$^*$ &76.3 \\
    RTFormer-B &RTX 2080Ti    &16.8M &$2048 \times 1024$ &- &39.1$^*$ &79.3 \\
    \hline
    SFNet-ResNet18  &RTX 3090 &12.3M  &$2048 \times 1024$ &50.5 &24.0 &79.0  \\
    PIDNet-S  &RTX 3090  &7.6M &$2048 \times 1024$ &- &93.2 &78.8 \\    
    PIDNet-M  &RTX 3090  &34.4M &$2048 \times 1024$ &- &39.8 &80.1 \\    
    \hline
    \textbf{SCTNet-S-Seg50}  &RTX 3090 &4.6M &$1024 \times 512$ &\textbf{451.2} &\textbf{160.3} &72.8\\
    \textbf{SCTNet-S-Seg75} &RTX 3090 &4.6M &$1536 \times 768$ &233.3 &149.2 &76.1 \\
    \textbf{SCTNet-B-Seg50}  &RTX 3090 &17.4M &$1024 \times 512$ &374.6 &144.9 &76.5 \\
    \textbf{SCTNet-B-Seg75}  &RTX 3090 &17.4M &$1536 \times 768$ &186.6 &105.2 &79.8\\
    \textbf{SCTNet-B-Seg100} &RTX 3090 &17.4M &$2048 \times 1024$ &105.0  &62.8 &\textbf{80.5} \\
    \Xhline{1pt}
  \end{tabular}
\end{table*}

\section{Disscussions}\label{section:D}
\subsection{Other Kinds of Real-time Semantic Segmentation Methods.} 
Our motivation is accelerating and improving the bilateral-branch networks for real-time semantic segmentation. So we focus on introducing the development of the bilateral-branch real-time semantic segmentation networks in related work, helping readers to understand our work in a limited space quickly. However, there are some other types of methods in that field. In order to avoid omissions, we discuss more comprehensive related work here.

\noindent
\textbf{Single-branch Networks.} There are some works focusing on single-branch real-time segmentation networks. However, lightweight single-branch networks usually lack long-range semantic information, leading to low accuracy. To capture high-quality semantic information and achieve high performance, AFFormer~\cite{dong2023afformer} use complex encoder blocks, SegNext~\cite{guo2022segnext} stacks lots of multi-scale convolution layers, SFNet~\cite{li2022sfnet} refers to a heavy decoder. All these results in larger inference latency, weakening the speed advantage of the original lightweight single-branch network.

\noindent
\textbf{Multi-branch Networks.} Multi-branch networks extend bilateral networks to more branches. For instance, PIDNet~\cite{xu2023pidnet} adopts a hierarchical main branch along with a detailed branch and a semantic branch. While multi-branch designs can enhance performance, they also come with higher computational costs, resulting in slower inference speeds compared to two-branch methods.

\subsection{Difference from MSCA.} 
There is also a convolutional attention in SegNext~\cite{guo2022segnext}, called MSCA(Multi-Scale Convolutional Attention). Although both the MSCA and the convolutional attention in our proposed CFBlock consist of several convolutions, there are very different. MSCA utilizes multi-scale stripe convolutions to capture multi-scale context, and then the extracted context feature map serves as the weight map to reweigh the origin feature map. Our convolutional attention utilizes external kernels as key and value, calculates the similarity between the external kernels and the feature map, and uses the value of external kernels to enhance the feature map. MSCA contains multi-scale convolution layers and larger kernels(like kernel size=21), resulting in more computational costs. As shown in Table 4(main paper), our CFBlock shows similar performance with fewer parameters and faster speed than MSCAN.

\begin{figure*}[ht!]
\small
\centering
\setlength{\fboxsep}{0pt}
\resizebox{1.0\linewidth}{!}{
\begin{tabular}{ccccc}
\vspace{-3pt}
\rotatebox{90}{~~~~~~~(a) Image} &
\includegraphics[width=0.19\linewidth]{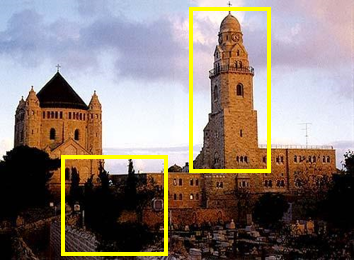}\label{ADE_1a} &
\hspace{-12pt}
\includegraphics[width=0.19\linewidth]{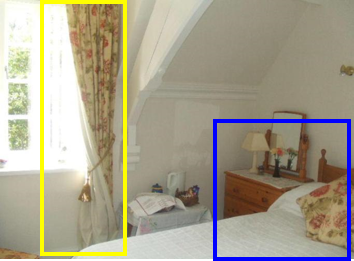}\label{ADE_2a} &
\hspace{-12pt}
\includegraphics[width=0.19\linewidth]{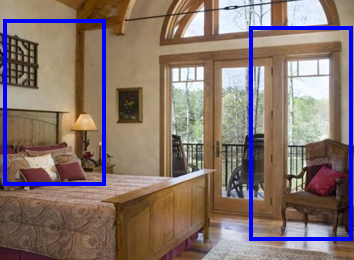}\label{ADE_3a} &
\hspace{-12pt}
\includegraphics[width=0.19\linewidth]{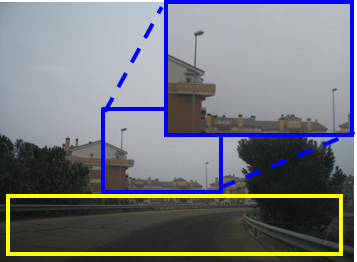}\label{ADE_4a} 
\\\vspace{-3pt}
\rotatebox{90}{~~~~~~~~~~(b) GT} &
\includegraphics[width=0.19\linewidth]{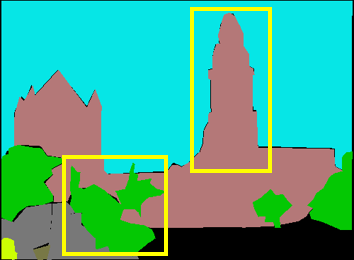}\label{ADE_1b} &
\hspace{-12pt}
\includegraphics[width=0.19\linewidth]{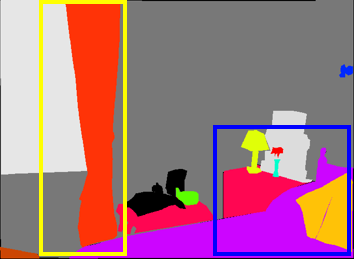}\label{ADE_2b} &
\hspace{-12pt}
\includegraphics[width=0.19\linewidth]{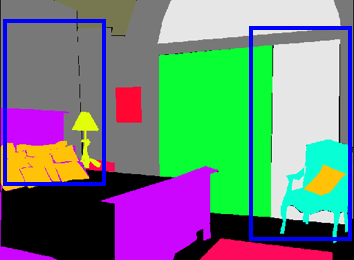}\label{ADE_3b} &
\hspace{-12pt}
\includegraphics[width=0.19\linewidth]{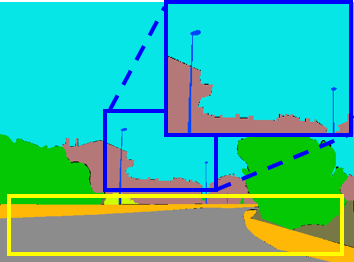}\label{ADE_4b} 
\\\vspace{-3pt}
\rotatebox{90}{(c) SegFormer-B0} &
\includegraphics[width=0.19\linewidth]{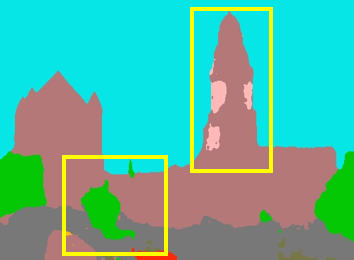}\label{ADE_1c} &
\hspace{-12pt}
\includegraphics[width=0.19\linewidth]{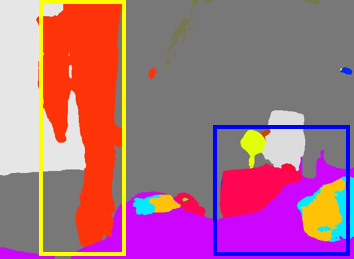}\label{ADE_2c} &
\hspace{-12pt}
\includegraphics[width=0.19\linewidth]{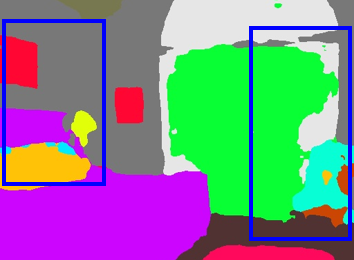}\label{ADE_3c} &
\hspace{-12pt}
\includegraphics[width=0.19\linewidth]{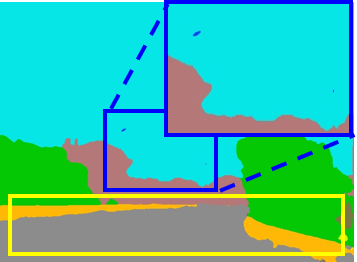}\label{ADE_4c} 
\\\vspace{-3pt}
\rotatebox{90}{(d) RTFormer-B} &
\includegraphics[width=0.19\linewidth]{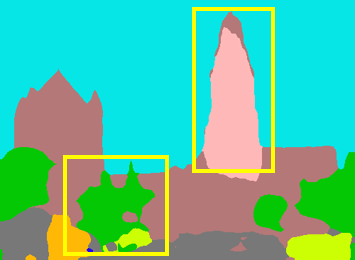}\label{ADE_1d} &
\hspace{-12pt}
\includegraphics[width=0.19\linewidth]{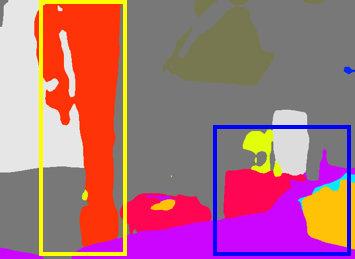}\label{ADE_2d} &
\hspace{-12pt}
\includegraphics[width=0.19\linewidth]{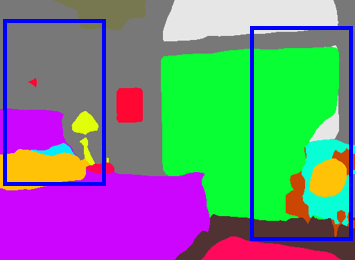}\label{ADE_3d} &
\hspace{-12pt}
\includegraphics[width=0.19\linewidth]{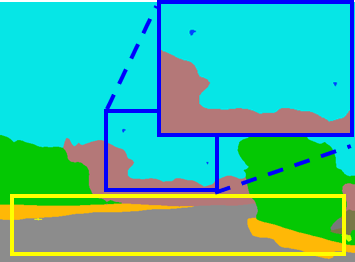}\label{ADE_4d} 
\\\vspace{-3pt}
\rotatebox{90}{(e) TopFormer-B} &
\includegraphics[width=0.19\linewidth]{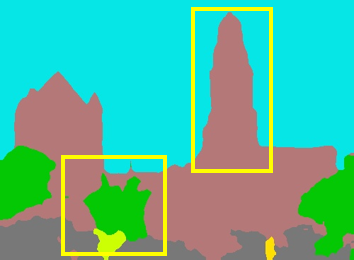}\label{ADE_1e} &
\hspace{-12pt}
\includegraphics[width=0.19\linewidth]{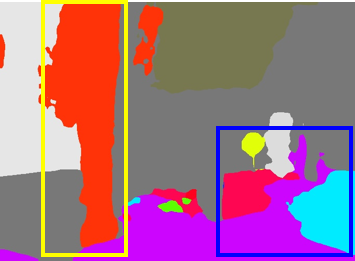}\label{ADE_2e} &
\hspace{-12pt}
\includegraphics[width=0.19\linewidth]{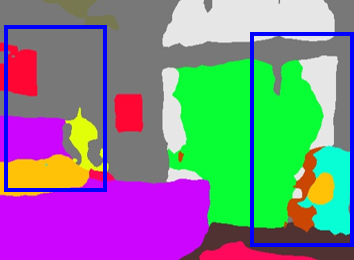}\label{ADE_3e} &
\hspace{-12pt}
\includegraphics[width=0.19\linewidth]{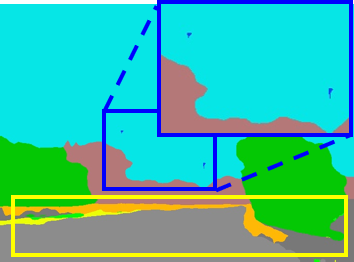}\label{ADE_4e} 
\\\vspace{-3pt}
\rotatebox{90}{(f) SeaFormer-B} &
\includegraphics[width=0.19\linewidth]{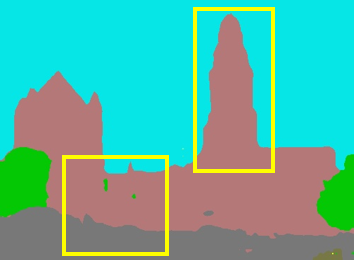}\label{ADE_1f} &
\hspace{-12pt}
\includegraphics[width=0.19\linewidth]{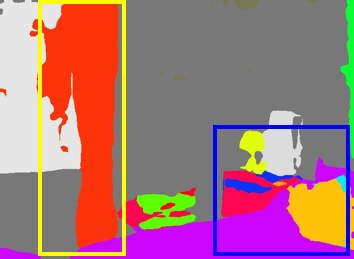}\label{ADE_2f} &
\hspace{-12pt}
\includegraphics[width=0.19\linewidth]{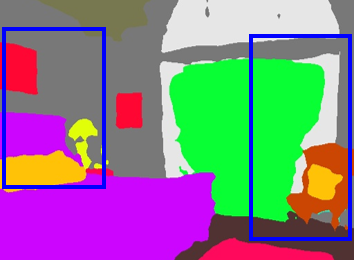}\label{ADE_3f} &
\hspace{-12pt}
\includegraphics[width=0.19\linewidth]{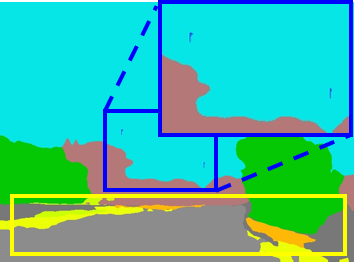}\label{ADE_4f} 
\\\vspace{-3pt}
\rotatebox{90}{(g) SCTNet-B} &
\includegraphics[width=0.19\linewidth]{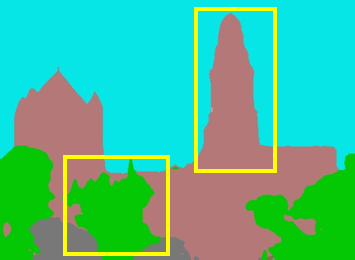}\label{ADE_1g} &
\hspace{-12pt}
\includegraphics[width=0.19\linewidth]{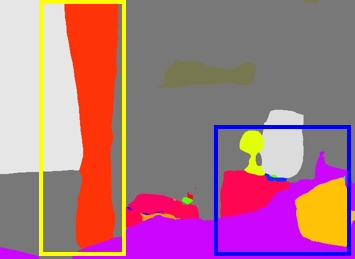}\label{ADE_2g} &
\hspace{-12pt}
\includegraphics[width=0.19\linewidth]{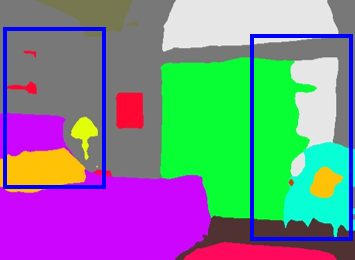}\label{ADE_3g} &
\hspace{-12pt}
\includegraphics[width=0.19\linewidth]{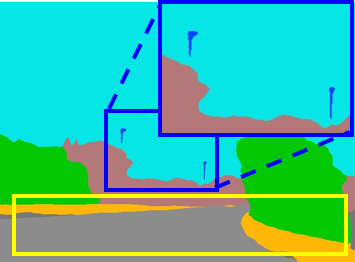}\label{ADE_4g} 
\\

 &  &  & &  \\
\end{tabular}

}

\caption{\textbf{Visualization results on ADE20K validation set.} Compared with SegFormer-B0~\cite{xie2021segformer}, RTFormer-B~\cite{wang2022rtformer}, TopFormer-B~\cite{zhang2022topformer} and SeaFormer-B\cite{wan2023seaformer}, SCTNet-B generates masks with finer details as highlighted in the dark blue box and more accurate large-area predictions, as highlighted in the yellow box.
    }
\label{suppfig:ADE_Visual}

\vspace{-10pt}
\end{figure*}

\begin{figure*}[t]
\centering
\setlength{\fboxsep}{0pt}
\begin{tabular}{ccccc}
\includegraphics[width=0.19\linewidth]{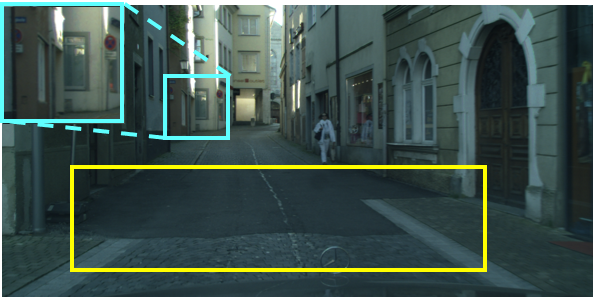}\label{city_1a} &
\hspace{-12pt}
\includegraphics[width=0.19\linewidth]{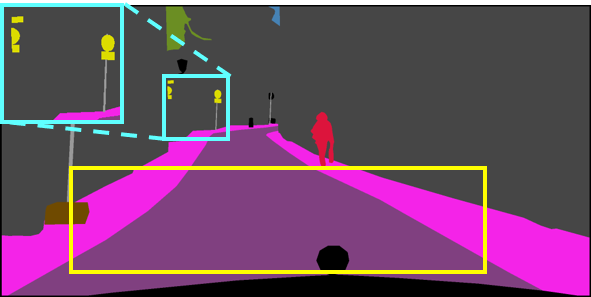}\label{city_1b} &
\hspace{-12pt}
\includegraphics[width=0.19\linewidth]{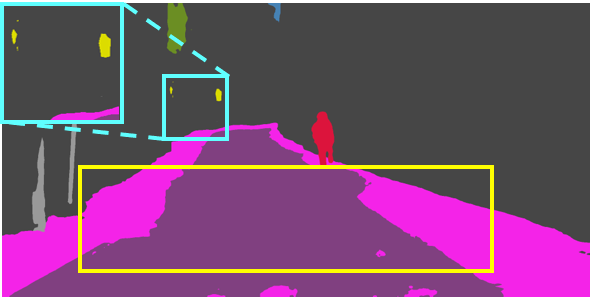}\label{city_1c} &
\hspace{-12pt}
\includegraphics[width=0.19\linewidth]{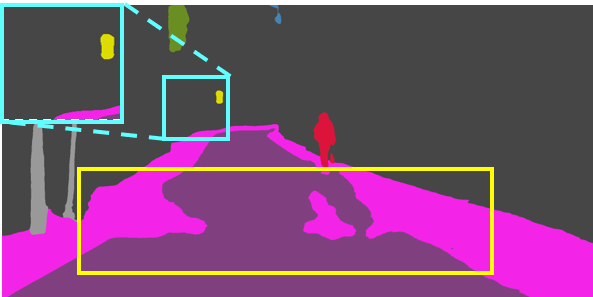}\label{city_1d} &
\hspace{-12pt}
\includegraphics[width=0.19\linewidth]{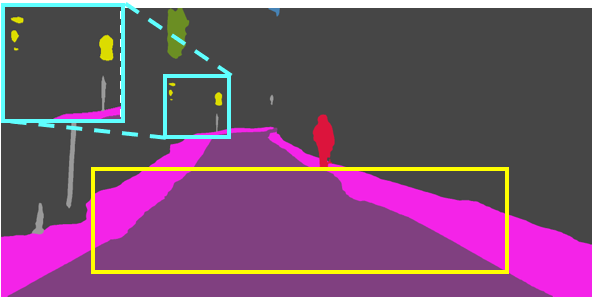}\label{city_1e} \\

\includegraphics[width=0.19\linewidth]{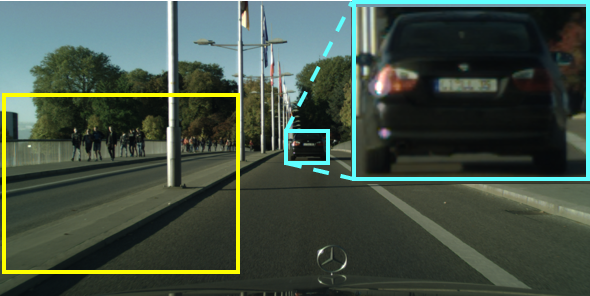}\label{city_2a} &
\hspace{-12pt}
\includegraphics[width=0.19\linewidth]{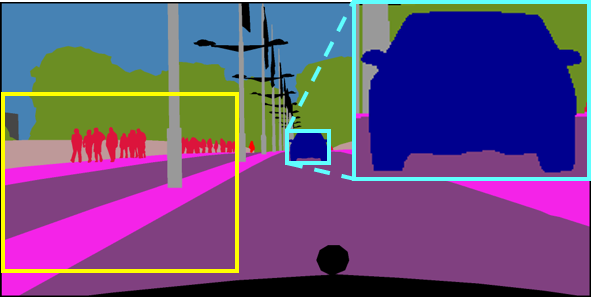}\label{city_2b} &
\hspace{-12pt}
\includegraphics[width=0.19\linewidth]{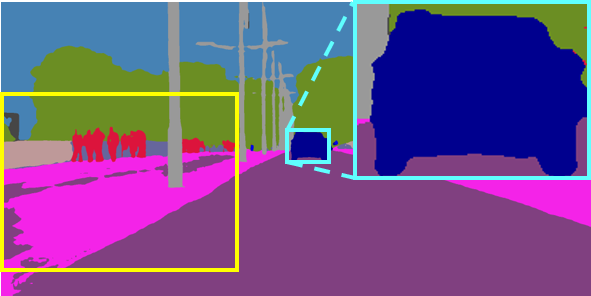}\label{city_2c} &
\hspace{-12pt}
\includegraphics[width=0.19\linewidth]{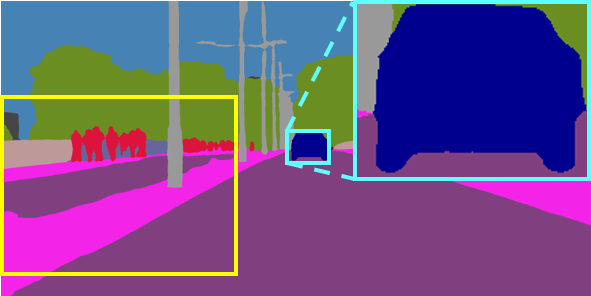}\label{city_2d} &
\hspace{-12pt}
\includegraphics[width=0.19\linewidth]{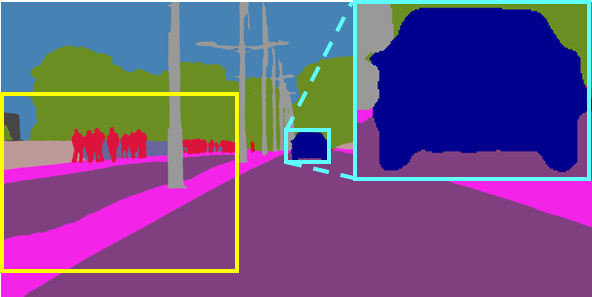}\label{city_2e} \\

\includegraphics[width=0.19\linewidth]{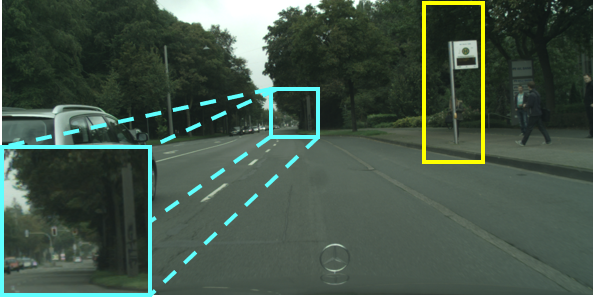}\label{city_3a} &
\hspace{-12pt}
\includegraphics[width=0.19\linewidth]{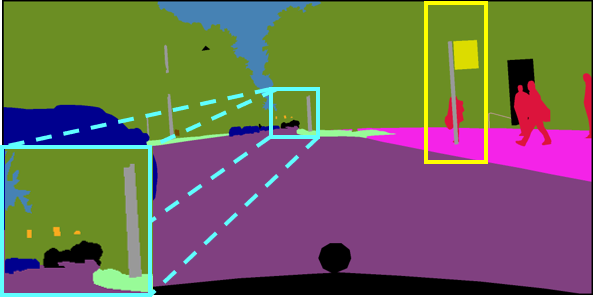}\label{city_3b} &
\hspace{-12pt}
\includegraphics[width=0.19\linewidth]{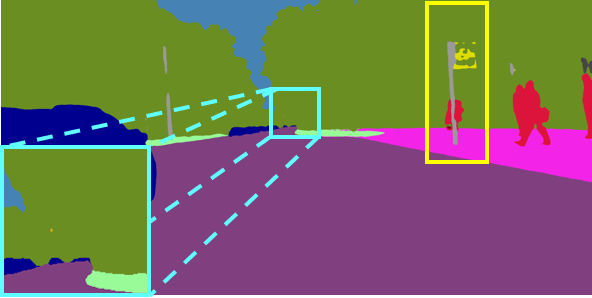}\label{city_3c} &
\hspace{-12pt}
\includegraphics[width=0.19\linewidth]{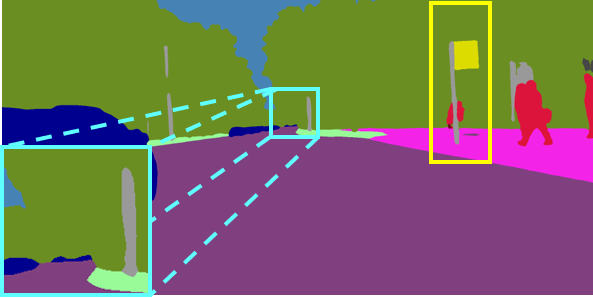}\label{city_3d} &
\hspace{-12pt}
\includegraphics[width=0.19\linewidth]{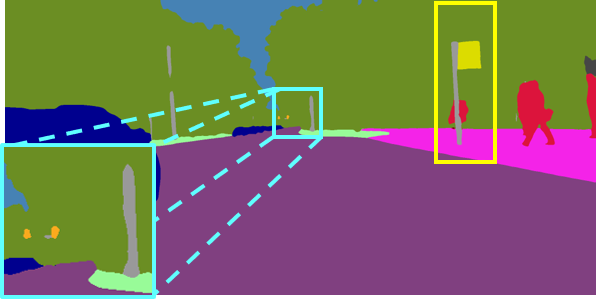}\label{city_3e} \\
(a) Image & (b) GT & (c) SeaFormer-B & (d) STDC2& (e) SCTNet-B\\
\end{tabular}
\caption{\textbf{Visualization results on Cityscapes validation set.} Compared with SeaFormer-B\cite{wan2023seaformer} and STDC2~\cite{fan2021rethinking}, SCTNet-B generates masks with finer details as highlighted in the light blue box and more accurate large-area predictions, as highlighted in the yellow box.}
\label{suppfig:vis_cityscapes}

\vspace{-10pt}
\end{figure*}

\subsection{Knowledge Distillation for Isomeric Models.} 
Knowledge distillation was first proposed by Hinton et al. ~\cite{hinton2015distilling}. Most distillation methods consider only homologous structures of teacher and student. Recently, some methods~\cite{touvron2021training,li2022locality} conduct distillation between CNN teacher and transformer student to tackle the problem of missing local information on small datasets for the transformer. Liu et al.~\cite{liu2022cross} studies distillation between transformer teacher and CNN student in classification. In the field of semantic segmentation, Zhu et al.~\cite{zhu2023good} introduced a collaborative learning process that improves the performance of CNN and transformer models. In contrast, our approach focuses on leveraging the high-quality semantic information of the transformer as a semantic branch to assist the learning of a single-branch CNN. As a result, we do not require joint training and have a smaller training cost. While our primary goal is to use the transformer to boost real-time semantic segmentation, the promising performance of our combined use of SIMA and CFBlock also provides insights for heterogeneous distillation from transformer to CNN.

\subsection{Limitations and Future work}
We demonstrate the effectiveness of our SCTNet on various GPU devices and datasets, both on torch and Tensor-RT models. However, we do not report CPU latency on mobile devices. Building a consistent benchmark for CPU latency and evaluating so many methods is beyond the scope of this paper. Additionally, the design of SCTNet does not include special modifications for CPU, such as replacing convolution operations with depth separable convolution or using ReLU6 instead of ReLU. We plan to modify SCTNet for CPU usage in future work.

While we have achieved a new state-of-the-art trade-off between accuracy and speed on various datasets, there is potential to further improve performance by scaling up our SCTNet to SCTNet-L. This is because our SCTNet-B exceeds the real-time standard(30FPS) by 32.8 FPS. We will explore this in future work and update our results in subsequent versions of this paper.

\section{More Visualization Results}\label{section:E}
In this section, we show more visualization results on ADE20K~\cite{zhou2017scene} and Cityscapes~\cite{cordts2016cityscapes} to demonstrate the effectiveness of the SCTNet intuitively. 

For ADE20K~\cite{zhou2017scene}, Figure~\ref{suppfig:ADE_Visual} presents the visualization results of all the mentioned transformer-based real-time segmentation methods and our SCTNet-B. In the first column, our SCTNet segments the buildings with better internal consistency and has a better mask for the tree. In the second column, our SCTNet has smooth boundaries of the curtain and the cushion like the GT masks. In the third column, our SCTNet generates better segmentation of the armchair and has fewer wrong masks for the painting. In the fourth column, our SCTNet generates better masks for fences and poles. These visualized results show that SCTNet has better capability to distinguish different classes when compared with SegFormer-B0~\cite{xie2021segformer}, RTFormer-B~\cite{wang2022rtformer}, TopFormer-B~\cite{zhang2022topformer} and SeaFormer-B\cite{wan2023seaformer}.

Moreover, we provide more visualization results on Cityscapes~\cite{cordts2016cityscapes} in Figure~\ref{suppfig:vis_cityscapes}, which are not present in the paper text because of the page limitation. Compared with SeaFormer~\cite{wan2023seaformer} and STDC~\cite{fan2021rethinking}, our SCTNet provides not only better results for those classes with large areas like road and sidewalk, but also more accurate boundaries for small or thin objects such as poles, traffic lights, traffic signs, and cars. This indicates that SCTNet extracts high-quality long-range context while preserving fine details.

\end{document}